\documentclass[]{youtu} 

\makeatletter
\renewcommand\tableofcontents{%
  \@starttoc{toc}%
}

\let\origaddcontentsline\addcontentsline
\newcommand{\DisableTOC}{\let\addcontentsline\@gobblethree}
\newcommand{\EnableTOC}{\let\addcontentsline\origaddcontentsline}
\makeatother

\usepackage{mathpazo}

\usepackage{graphicx}
\usepackage[numbers]{natbib}
\usepackage{float}
\usepackage{placeins}
\usepackage{wrapfig}
\usepackage{enumitem}
\usepackage[most]{tcolorbox}
\usepackage{subcaption}
\usepackage[dvipsnames]{xcolor}
\usepackage{fontawesome}
\usepackage{multirow}
\usepackage{amsmath}
\usepackage{amssymb}
\usepackage{algorithm}
\usepackage{algorithmic}
\usepackage{newfloat}
\usepackage{listings}
\usepackage{booktabs}
\usepackage{adjustbox}
\usepackage{array}
\usepackage{tabularx}
\usepackage{needspace}

\DeclareCaptionStyle{ruled}{labelfont=normalfont,labelsep=colon,strut=off}
\floatstyle{ruled}
\newfloat{listing}{tb}{lst}{}
\floatname{listing}{Listing}

\definecolor{myborder}{RGB}{73,86,102}
\definecolor{myRed}{RGB}{240,48,159}
\definecolor{mylightblue}{RGB}{235,245,255}
\tcbuselibrary{skins}

\title{SCOPE-Router: Cost-Aware Open-Set VLM Routing for Execution-Oriented Tasks}
\author{
Tao Yu\textsuperscript{$1,2 *$},
Yifei Qu\textsuperscript{$1 * \dag$},
Zhiqing Cui\textsuperscript{$1 * \dag$},
Pengfei Zhou\textsuperscript{$3 \ddagger$},
Zhongtian Luo\textsuperscript{$1, \dag$},
Yujia Yang\textsuperscript{$2$},
Shenghua Chai\textsuperscript{$1, \dag$},
Haopeng Jin\textsuperscript{$4$},
Zhenghao Zhang\textsuperscript{$2$},
Xinming Wang\textsuperscript{$1,2$},
Hongzhu Yi\textsuperscript{$2 \ddagger$},
Wangbo Zhao\textsuperscript{$3$},
Zhenglin Wan\textsuperscript{$3$},
Yan Huang\textsuperscript{$1,2 \ddagger$},
Yeshani\textsuperscript{$4$},
Jinwen Luo\textsuperscript{$4$},
Yang You\textsuperscript{$3$}
}

\affiliation{
\textsuperscript{$1$}CASIA \quad
\textsuperscript{$2$}UCAS \quad
\textsuperscript{$3$}NUS \quad
\textsuperscript{$4$}Tencent 

}

\date{Aug. 19, 2026}
\sourcecode{https://github.com/yutao1024/SCOPE-Router}
\data{https://huggingface.co/datasets/Kirito-Lab/VLM-ExecRouterBench}

\abstract{Model routing aims to select the most suitable model from a candidate pool for each query, balancing quality and cost. Existing VLM routing research is limited to traditional VQA evaluation, lacks systematic calibration optimization for open-set scenarios, and employs training objectives that dilute multi-positive signals via softmax normalization without incorporating cost. We address these limitations with three contributions: (1)~VLM-ExecRouterBench, the first execution-oriented VLM routing benchmark covering Code, Agentic, and Search domains with 11 candidate models spanning nearly two orders of magnitude in pricing; (2)~SCOPE-Router, a dual-tower router that matches queries to model behavior profiles constructed via hybrid calibration (random/diagnostic/diversity sampling), enabling new models to join routing without retraining; (3)~CRM+RCCR, an architecture-agnostic cost-aware objective that encodes cost preference into continuous relevance targets through per-pair independent scoring, eliminating multi-positive dilution while regularizing queries with similar routing preferences to be closer in the routing space. Empirically, SCOPE-Router achieves the best Rank Score on all three benchmarks, surpassing the runner-up by 1.84 points under OOD settings and by 6.75 points under doubly OOD open-set evaluation. When applied to four diverse routers, CRM+RCCR improves Rank Score by 1.25--6.21 points.}

\begin{document}
\maketitle

\begin{NoHyper}
\renewcommand{\thefootnote}{*}
\footnotetext{Equal contribution.}
\renewcommand{\thefootnote}{\dag}
\footnotetext{Work done during an internship at CASIA.}
\renewcommand{\thefootnote}{\textdaggerdbl}
\footnotetext{Corresponding author.}
\end{NoHyper}
\renewcommand{\thefootnote}{\arabic{footnote}}

\DisableTOC

\section{Introduction}


The rapid advancement of LLMs and VLMs has given rise to a heterogeneous model ecosystem. From open-source models with fewer than 8B parameters to flagship commercial APIs, models differ substantially in capability, cost, and latency. No single model achieves optimal accuracy at minimal cost across all tasks~\cite{routerbench,vlrouterbench,routerarena}, motivating model routing: selecting the most suitable model per query to balance quality and cost~\cite{routellm,carrot,zhou2026agent}. In practice, VLMs are increasingly deployed not through raw API calls for simple question answering, but within execution pipelines, including code generation, tool-calling agents, and multi-step web retrieval, where routing decisions directly impact both task success and deployment cost.

Recent years have seen progress in LLM/VLM routing. Methodologically, existing work spans classifier-based predictive routing, contrastive learning matching~\cite{routerdc}, cluster-based generalization~\cite{uniroute}, and reinforcement learning orchestration~\cite{xrouter}. On the evaluation side, VL-RouterBench first extends routing evaluation to vision-language models, covering 14 datasets and 17 candidate VLMs; MMR-Bench~\cite{mmrbench} further introduces budget-aware multimodal reasoning evaluation. However, existing work faces limitations along three dimensions:

\textbf{Limited domain coverage.} Current VLM routing benchmarks cover only traditional VQA tasks---visual reasoning, OCR, chart understanding---where models directly output text answers. Yet VLMs are increasingly applied to tasks that require active execution: code generation demands precise program logic verified by unit tests, tool-calling tasks require multi-step reasoning with external tool interaction, and web search involves planning retrieval strategies and synthesizing multi-source information. These execution tasks impose different capability requirements and exhibit stronger inter-model complementarity, yet no routing benchmark incorporates them.

\textbf{Insufficient open-set capability.} New models emerge continuously in practice, but traditional routing methods treat models as fixed class labels, requiring retraining upon each addition. Existing open-set approaches (UniRoute, ICL-Router~\cite{iclrouter}) lack systematic optimization of calibration data selection: UniRoute randomly partitions validation data for model representation, while ICL-Router filters samples by a single signal. Under limited calibration budgets, the absence of targeted selection strategies leads to low discriminability in model profiles.

\begin{figure*}[t]
    \centering
    \includegraphics[width=0.95\linewidth]{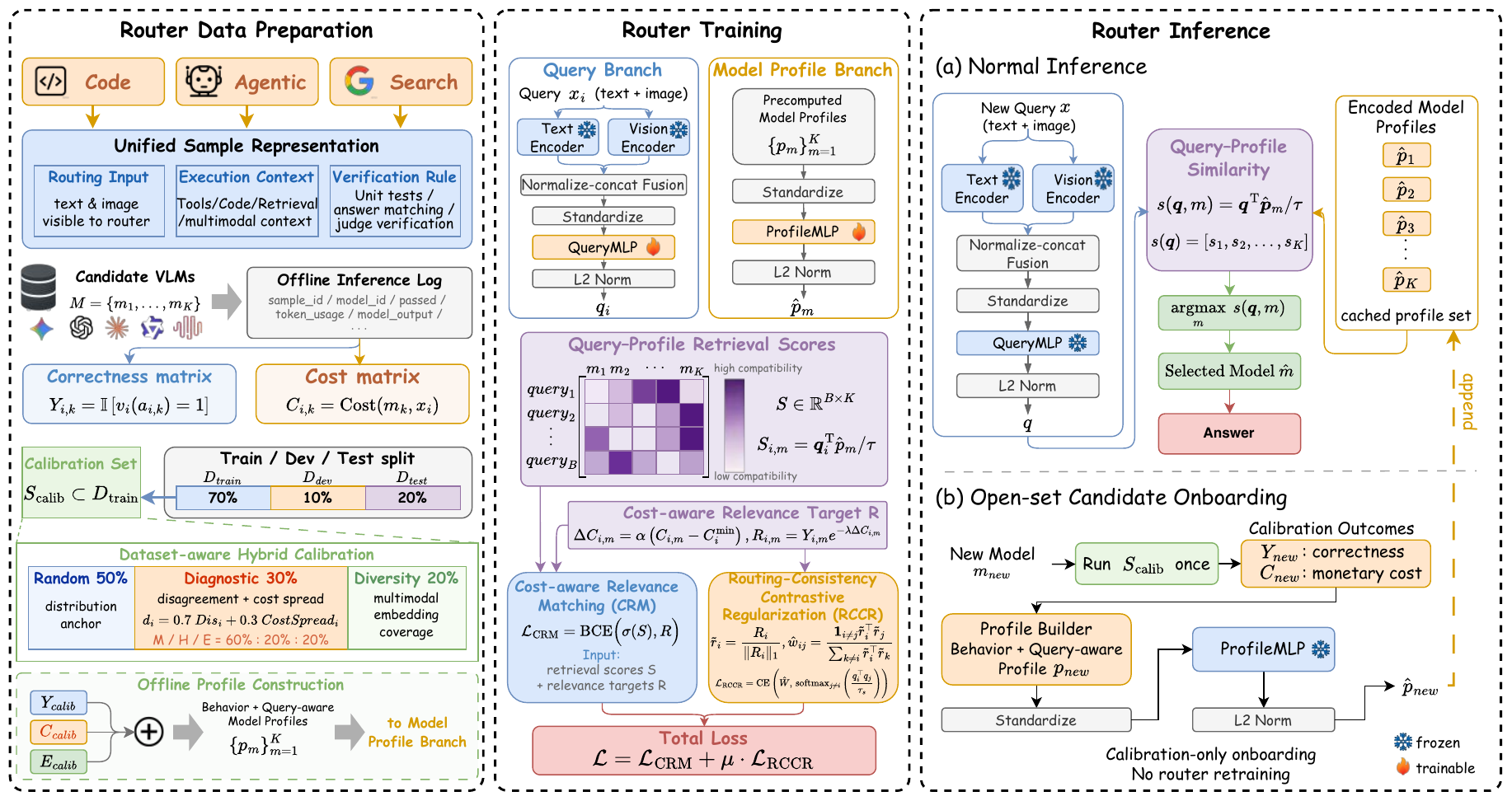}
    \caption{
    Overview of the proposed SCOPE-Router framework.
    Left: router data preparation constructs unified execution samples,
    correctness and cost matrices, and query-aware model profiles using the
    hybrid calibration set.
    Middle: during router training, the query and model-profile branches are
    projected into a shared routing space and jointly optimized using
    Cost-aware Relevance Matching (CRM) and Routing-Consistency Contrastive
    Regularization (RCCR).
    Right: during inference, the router selects the most suitable candidate
    according to query--profile similarity; a new model can be incorporated
    by running only the calibration set to construct and cache its profile,
    without retraining the router.
    }
    \label{fig:framework}
\end{figure*}

\textbf{Training objectives decoupled from cost.} RouterDC's contrastive objective partitions positive/negative examples based solely on correctness, ignoring cost differences; VL-RouterBench's softmax CE loss dilutes multi-positive signals through normalization; UniRoute only incorporates cost as a post-hoc inference penalty, leaving the training process itself cost-unaware.

To address these limitations, we make the following contributions (see Figure~\ref{fig:framework} for an overview):

\textbf{(1) VLM-ExecRouterBench: the first execution-oriented VLM routing benchmark.} We extend evaluation from traditional VQA to three execution domains: code generation (Code), tool-calling agents (Agentic), and multi-step web search (Search). Each sample is standardized into a Routing Input / Execution Context / Verification Rule triplet, uniformly covering three distinct execution protocols. The candidate pool comprises 11 models from five major providers, with output pricing spanning nearly two orders of magnitude.

\textbf{(2) SCOPE-Router: an open-set dual-tower router with hybrid calibration.} We formulate open-set routing as query-to-profile matching: for each model, we construct a Query-Aware Profile that fuses behavioral statistics and semantic directions, and compute matching scores via lightweight dual-tower MLPs in a shared routing space. New models need only run once on a small calibration set to generate a profile and immediately join routing, without retraining. To maximize profile discriminability under limited budgets, we design a hybrid calibration strategy allocating budgets at 50\%/30\%/20\% to random sampling (distributional coverage), diagnostic sampling (driven by model disagreement and cost spread), and diversity sampling (uniform semantic coverage). The three strategies are complementary, with the full combination outperforming any single strategy.

\textbf{(3) CRM+RCCR cost-aware training objective.} We propose \textbf{C}ost-aware \textbf{R}elevance \textbf{M}atching (CRM), which defines a continuous relevance target that encodes cost preference directly into supervision and adopts per-pair independent sigmoid binary cross-entropy (BCE) in place of softmax cross-entropy (CE) to avoid multi-positive signal dilution. This is complemented by \textbf{R}outing-\textbf{C}onsistency \textbf{C}ontrastive \textbf{R}egularization (RCCR), which pulls queries with similar routing patterns closer in the routing space, enabling the router to generalize across queries that share model suitability. The objective is architecture-agnostic and can directly replace existing router losses. Extensive experiments on RouterDC, ZOOTER~\cite{zooter}, CosineCls, and VLC all show consistent Rank Score improvements without any structural modification.

We conduct systematic experiments on VLM-ExecRouterBench, VL-RouterBench, and MMR-Bench. SCOPE-Router achieves the best Rank Score on all three benchmarks and maintains its lead under OOD generalization settings and open-set evaluation. Ablation studies verify the complementarity of CRM and RCCR, the effectiveness of the hybrid calibration strategy, and the cross-architecture transferability of the proposed training objective.

\section{Related Work}
\label{sec:related}

\paragraph{LLM Routing Benchmarks.}
RouterBench formulated LLM routing as cost--quality optimization. LLMRouterBench~\cite{llmrouterbench} and RouterArena further expanded model pools and evaluation dimensions for text routing. VL-RouterBench extended the paradigm to VLMs with 14 datasets, 17 models, and a unified Rank Score metric. However, existing VLM benchmarks remain confined to traditional VQA tasks. Execution-based tasks such as code generation, tool-use agentic reasoning, and web retrieval exhibit stronger inter-model complementarity yet have not been incorporated into any routing benchmark. Our VLM-ExecRouterBench is the first to cover these domains.

\paragraph{Model Representation and Open-Set Routing.}
Early routers map queries to fixed model IDs, requiring retraining for new models. Recent open-set approaches use per-cluster error rates, anchor pass/fail patterns, or public-signal GNNs~\cite{routeprofile}, but each sacrifices sample-level granularity, cost information, or behavioral fidelity. Our profiles are derived from first-hand inference on a carefully designed hybrid calibration set, fusing per-sample correctness, cost, and semantic directions to capture both what a model can solve and at what price.

\paragraph{Router Training Objectives.}
RouterDC uses dual contrastive losses but partitions positives purely by correctness. VL-RouterBench pairs cost-weighted soft labels with softmax CE, but normalization dilutes gradients when multiple models answer correctly. UniRoute treats cost only at inference time. We propose CRM+RCCR: Cost-aware Relevance Matching encodes cost preference into continuous relevance targets and eliminates multi-positive dilution, while Routing-Consistency Contrastive Regularization pulls queries with similar routing patterns closer in the routing space, enabling the router to benefit from shared model-suitability structure.

\section{Method}
\label{sec:method}

\subsection{Problem Formulation}
\label{sec:problem}

Model routing comprises three stages: data construction, router training, and routing inference.

\paragraph{Data Construction.}
Given task queries $\{x_i\}_{i=1}^N$ (text and/or images) and a candidate VLM pool $\mathcal{M} = \{m_1, \ldots, m_K\}$, we execute every model on every sample, yielding a correctness matrix $Y \in \{0,1\}^{N \times K}$ ($Y_{i,m}=1$ if model~$m$ answers sample~$i$ correctly) and a cost matrix $C \in \mathbb{R}_+^{N \times K}$ ($C_{i,m}$ is the inference cost computed from token usage and per-million-token pricing).

\paragraph{Router Training.}
Given $Y$ and $C$, the router learns a scoring function $s_\theta(x, m)\colon \mathcal{X} \times \mathcal{M} \to \mathbb{R}$. High-scoring models should be both likely correct and relatively cheap. The training objective must encode quality and cost preferences.

\paragraph{Routing Inference.}
Given a new query $x_i$, the router selects $\hat{m}_i = \arg\max_{m_k \in \mathcal{M}} s_\theta(x_i, m_k)$. Evaluation uses the \emph{Rank Score}:
\begin{equation}
\label{eq:rank_score}
S_\beta = \frac{(1+\beta) \cdot A \cdot \widetilde{C}}{\beta \cdot A + \widetilde{C}},
\end{equation}
where $A$ is routing accuracy, $\widetilde{C} = \frac{\log_2 c_{\max} - \log_2 \mathrm{cost}}{\log_2 c_{\max} - \log_2 c_{\min}}$ is the log-normalized cost score, and $\beta$ weights accuracy vs.\ cost (details in the Evaluation Metrics paragraph).

\subsection{Data Generation}
\label{sec:data}

Existing VLM routing benchmarks draw exclusively from traditional VQA-style datasets. We extend evaluation to execution-oriented domains (Code, Agentic, Search) by generating standardized samples from existing domain-specific sources 
(see the Datasets section for detailed dataset descriptions and Appendix~\ref{app:data_generation} for data-generation procedures and implementation details).

Each sample is converted into a three-part structure: (1)~\textbf{Routing Input}: the query visible to the router; (2)~\textbf{Execution Context}: the full context given to each candidate VLM; (3)~\textbf{Verification Rule}: the validation logic that judges model output. This design separates what the router sees from what the model executes.

\paragraph{Matrix Construction.}
For each generated sample $x_i$, all models execute the Execution Context and the Verification Rule judges the results, producing the correctness matrix $Y$ and cost matrix $C$.

\paragraph{Data Splits.}
The full dataset is partitioned 70\%/10\%/20\%: training, validation, and held-out test.

\subsection{Open-Set Routing and Hybrid Calibration Sets}
\label{sec:openset}

Open-set routing enables new models to integrate without retraining. Instead of treating models as fixed class labels, we build a \emph{behavioral profile vector} for each model. The router learns to match queries to profiles---a new model need only be executed once on a small calibration set to generate its profile and immediately participate in routing.

\paragraph{Hybrid Calibration Set Selection.}
\label{sec:calibration_selection}

The calibration set $\mathcal{S}_{\mathrm{calib}} \subset \mathcal{S}_{\mathrm{train}}$ ($|\mathcal{S}_{\mathrm{calib}}| = 1024$) is selected via a hybrid strategy allocating the budget at a 50\%/30\%/20\% ratio across three complementary mechanisms, executed sequentially.

\paragraph{Random Sampling (50\%).}
Uniform sampling without replacement, ensuring unbiased coverage of the training distribution.

\paragraph{Diagnostic Sampling (30\%).}
From remaining samples, this component maximizes discriminative power. For each candidate sample $i$:
\begin{equation}
\label{eq:diagnostic}
d_i = w_{\mathrm{dis}} \cdot \underbrace{4\,p_i(1-p_i)}_{\text{disagreement}} + w_{\mathrm{cost}} \cdot \underbrace{\mathrm{cost\_spread}_i^{\,\mathrm{norm}}}_{\text{cost spread}},
\end{equation}
where $p_i = \frac{1}{K}\sum_{m=1}^{K} Y_{i,m}$ is the fraction of models in the reference pool $\mathcal{M}$ answering correctly ($w_{\mathrm{dis}}=0.7$, $w_{\mathrm{cost}}=0.3$). The disagreement term peaks at $p_i=0.5$ (maximum inter-model disagreement) and equals zero when all models agree. The cost spread is defined as the difference between the maximum and minimum inference cost among models that answer correctly, normalized by the 95th percentile across all samples for robustness. Samples are stratified into difficulty buckets (hard $p_i \leq 0.25$ / medium $(0.25, 0.75]$ / easy $> 0.75$, allocated at 20\%/60\%/20\%) and drawn with weights proportional to $\exp(\hat{d}_i / \tau_{\mathrm{samp}})$ within each bucket, where $\hat{d}_i$ is the min-max normalized score within each bucket ($\tau_{\mathrm{samp}}=0.6$).

\paragraph{Diversity Sampling (20\%).}
From the remaining unselected samples, we apply MiniBatchKMeans clustering on their pre-computed query embeddings (text and vision encoder features fused via normalize-concat), with the number of clusters set equal to the target sample count. The sample closest to each cluster centroid is selected, ensuring uniform coverage of the embedding space. Empty clusters (rare) are filled by selecting the farthest points from the global centroid.

\paragraph{Model Behavior Profile Construction.}
\label{sec:profile}

Given the calibration set ($S = 1024$ samples), we construct a \emph{Query-Aware Profile} $\mathbf{p}_m$ for each model~$m$.

\paragraph{Behavioral Vector.}
\begin{equation}
\label{eq:behavior}
\mathbf{p}_m^{\mathrm{behav}} = [\,\mathbf{y}_m;\; \tilde{\mathbf{c}}_m;\; \mathbf{y}_m \odot (1-\tilde{\mathbf{c}}_m);\; \mathbf{s}_m\,],
\end{equation}
where $\mathbf{y}_m \in \{0,1\}^S$ is per-sample correctness, $\tilde{\mathbf{c}}_m \in [0,1]^S$ is min-max normalized cost, the third term is the ``value'' vector (high weight for samples answered correctly at low cost), and $\mathbf{s}_m \in \mathbb{R}^8$ is a summary statistics vector containing accuracy, mean cost (raw and normalized), mean cost on correct/incorrect subsets, mean value score, and correct/incorrect fractions.

\paragraph{Semantic Vector.}
Query embeddings from the calibration set are L2-normalized, then aggregated by model performance:
\begin{equation}
\label{eq:semantic}
\mathbf{p}_m^{\mathrm{sem}} = [\,\bar{\mathbf{e}}_m^{+};\; \bar{\mathbf{e}}_m^{-};\; \bar{\mathbf{e}}_m^{v}\,],
\end{equation}
where
\begin{gather}
\bar{\mathbf{e}}_m^{+} = \frac{\sum_{i} Y_{i,m} \cdot \hat{\mathbf{e}}_i}{\sum_{i} Y_{i,m}}, \quad
\bar{\mathbf{e}}_m^{-} = \frac{\sum_{i} (1{-}Y_{i,m}) \cdot \hat{\mathbf{e}}_i}{\sum_{i} (1{-}Y_{i,m})}, \notag\\
\bar{\mathbf{e}}_m^{v} = \frac{\sum_{i} Y_{i,m}(1{-}\tilde{c}_{i,m}) \cdot \hat{\mathbf{e}}_i}{\sum_{i} Y_{i,m}(1{-}\tilde{c}_{i,m})},
\end{gather}
with $\hat{\mathbf{e}}_i = \mathbf{e}_i / \|\mathbf{e}_i\|_2$. The three directional vectors encode what the model is good at, bad at, and efficient at in semantic space.

The final profile is $\mathbf{p}_m = [\mathbf{p}_m^{\mathrm{behav}};\; \mathbf{p}_m^{\mathrm{sem}}] \in \mathbb{R}^{3S+8+3D}$, where $D$ is the query embedding dimension (with default encoders BGE-M3~\cite{bge-m3} + DINOv2-large`\cite{oquab2023dinov2}, $D = 2048$).

\subsection{Router Architecture}
\label{sec:architecture}

We propose \textbf{SCOPE-Router} (\textbf{S}elective \textbf{C}alibration for \textbf{O}pen-set \textbf{P}rofile-\textbf{E}nhanced Routing)---a dual-tower architecture projecting queries and model profiles into a shared routing space.

\paragraph{Query Encoder.}
Query $x_i$ is encoded via text and vision encoders with normalize-concat fusion to form $\mathbf{e}_i$. After standardization, a lightweight QueryMLP projects to the routing space:
\begin{equation}
\label{eq:query_enc}
\mathbf{q}_i = \ell_2\bigl(\mathrm{QueryMLP}(\mathrm{standardize}(\mathbf{e}_i))\bigr),
\end{equation}
where $\ell_2(\mathbf{v}) = \mathbf{v} / \|\mathbf{v}\|_2$ denotes L2 normalization.

\paragraph{Profile Encoder.}
Model profiles $\mathbf{p}_m$ are likewise standardized and projected via ProfileMLP:
\begin{equation}
\label{eq:profile_enc}
\hat{\mathbf{p}}_m = \ell_2\bigl(\mathrm{ProfileMLP}(\mathrm{standardize}(\mathbf{p}_m))\bigr).
\end{equation}

Both MLPs are 2-layer (hidden dimension 128 with ReLU and dropout; output dimension 64). L2-normalized outputs produce the matching score:
\begin{equation}
\label{eq:score}
s_{i,m} = \mathbf{q}_i^\top \hat{\mathbf{p}}_m \,/\, \tau,
\end{equation}
where $\tau$ is a temperature parameter. At inference, the router selects $\arg\max_m\, s_{i,m}$.

\subsection{Cost-Aware Training Objective}
\label{sec:loss}

Standard routing training faces two problems: (1)~hard-label softmax discards information about other correct models; (2)~cost preference is only applied post-hoc. We propose a unified solution.

\paragraph{Relevance Target.}
For each query--model pair:
\begin{equation}
\label{eq:relevance}
R_{i,m} = \mathbb{1}[Y_{i,m}=1] \cdot \exp\bigl(-\lambda \cdot \alpha \cdot (C_{i,m} - C_i^{\min})\bigr),
\end{equation}
where $C_i^{\min} = \min_{m':Y_{i,m'}=1} C_{i,m'}$ is the cost of the cheapest correct model. The cheapest correct model receives $R=1$, more expensive correct models receive $R \in (0,1)$, and incorrect models receive $R=0$. For all-failed samples ($\sum_m Y_{i,m}=0$), we set $R_{i,:}=\mathbf{0}$; they contribute to CRM as all-negative examples but are excluded from RCCR via $\mathcal{U}$. Edge cases in diagnostic sampling and profile aggregation (empty correct/incorrect sets) default to zero.

\paragraph{CRM Loss.}
We predict each pair independently via binary cross-entropy rather than competing models via row-wise softmax:
\begin{equation}
\label{eq:crm}
\mathcal{L}_{\mathrm{CRM}} = \frac{1}{BK} \sum_{i=1}^{B}\sum_{m=1}^{K} \mathrm{BCE}\bigl(\sigma(s_{i,m}),\; R_{i,m}\bigr),
\end{equation}
where $\sigma(\cdot)$ denotes the sigmoid function. Each pair's gradient is independent, so positive signals are not diluted in multi-positive scenarios.

\paragraph{Routing-Consistency Contrastive Regularization (RCCR).}
Let $\tilde{\mathbf{r}}_i = R_{i,:} / \|R_{i,:}\|_1$ be the L1-normalized relevance vector. We compute pairwise sample similarity $w_{ij} = \tilde{\mathbf{r}}_i^\top \tilde{\mathbf{r}}_j$ and row-normalize into $\hat{w}_{ij}$. Define $\mathcal{U} = \{i : \|R_{i,:}\|_1 > 0\}$ and $\mathcal{V} = \{i \in \mathcal{U} : \sum_{j \ne i} w_{ij} > 0\}$:
\begin{equation}
\label{eq:rccr}
\mathcal{L}_{\mathrm{RCCR}} = -\frac{1}{|\mathcal{V}|}\sum_{i \in \mathcal{V}} \sum_{\substack{j \in \mathcal{U} \\ j \ne i}} \hat{w}_{ij} \log \frac{\exp(\mathbf{q}_i^\top \mathbf{q}_j / \tau_s)}{\displaystyle\sum_{\ell \in \mathcal{U},\, \ell \ne i}\exp(\mathbf{q}_i^\top \mathbf{q}_\ell / \tau_s)},
\end{equation}
where $\tau_s$ is a temperature parameter.

\paragraph{Total Loss.}
\begin{equation}
\label{eq:total_loss}
\mathcal{L} = \mathcal{L}_{\mathrm{CRM}} + \mu \cdot \mathcal{L}_{\mathrm{RCCR}},
\end{equation}
with $\mu = 0.1$ by default. Since this objective operates solely on $Y$, $C$, and the router's output logits, it is architecture-agnostic and can be applied to any trainable router as a replacement loss.


\Needspace{12\baselineskip}
\section{VLM-ExecRouterBench}
\label{sec:benchmark}

\subsection{Datasets}
\label{sec:datasets}

\noindent
\begin{minipage}[t]{0.53\textwidth}
\vspace{0pt}

Our routing benchmark spans three task domains with 34k samples
(374k sample-model pairs) and 5.1B input-output tokens.
All datasets use the unified Routing Input / Execution Context /
Verification Rule structure (described in the Data Generation section),
split 70\%/10\%/20\%.

\noindent\textbf{Code.}
MBPP~\cite{mbpp}, BigCodeBench~\cite{bigcodebench},
APPS~\cite{apps}, and LiveCodeBench~\cite{livecodebench}
cover code synthesis from introductory to competition-level,
verified by unit tests.

\noindent\textbf{Agentic.}
MathVista~\cite{mathvista}, ChartQA~\cite{chartqa},
MMMU~\cite{mmmu}, OCRBench~\cite{ocrbench},
DocVQA~\cite{docvqa}, AI2D~\cite{ai2d},
and RealWorldQA~\cite{realworldqa}
are reformulated under an agentic setting requiring multi-step tool calls.

\noindent\textbf{Search.}
BrowseComp-Plus~\cite{browsecompplus}
is a deep-research benchmark derived from BrowseComp with a fixed,
human-verified corpus. Models perform multi-step retrieval over the
provided document collection to answer complex queries.

\end{minipage}
\hfill
\begin{minipage}[t]{0.43\textwidth}
\vspace{0pt}
\centering

\includegraphics[width=\linewidth]{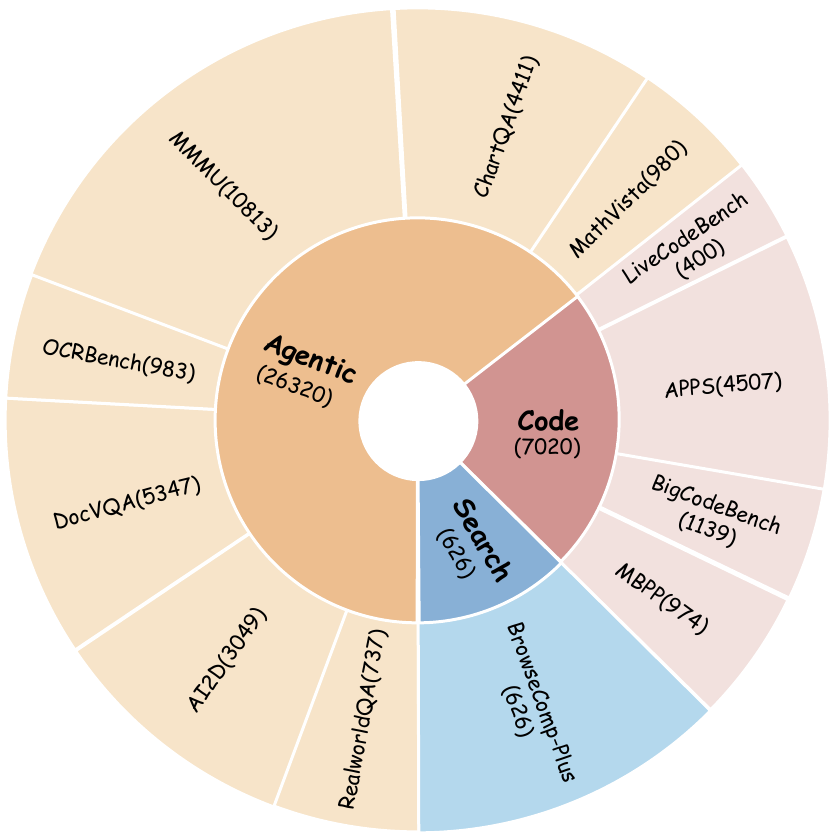}
\captionsetup{
    justification=centering,
    singlelinecheck=false
}
\captionof{figure}{Data composition of VLM-ExecRouterBench.}
\label{fig:data}

\end{minipage}

\par
\vspace{0.25em}

\noindent\textbf{Candidate Model Pool.}
\phantomsection
\label{sec:model_pool}

\begin{center}
\begin{minipage}{0.70\textwidth}
\centering

\resizebox{\linewidth}{!}{%
\begin{tabular}{clccc}
\toprule
\# & Model & Provider & Input (\$) & Output (\$) \\
\midrule
1  & Qwen3-VL-8B-Instruct   & Qwen / Alibaba & 0.08 & 0.50 \\
2  & Gemini 2.5 Flash Lite  & Google         & 0.10 & 0.40 \\
3  & Qwen3.5-35B-A3B        & Qwen / Alibaba & 0.14 & 1.00 \\
4  & MiniMax M3             & MiniMax        & 0.30 & 1.20 \\
5  & Gemini 3 Flash Preview & Google         & 0.50 & 3.00 \\
6  & GPT-5.4 mini           & OpenAI         & 0.75 & 4.50 \\
7  & Claude Haiku 4.5       & Anthropic      & 1.00 & 5.00 \\
8  & Gemini 3.5 Flash       & Google         & 1.50 & 9.00 \\
9  & GPT-5.4                & OpenAI         & 2.50 & 15.00 \\
10 & Claude Sonnet 4.6      & Anthropic      & 3.00 & 15.00 \\
11 & Claude Opus 4.6        & Anthropic      & 5.00 & 25.00 \\
\bottomrule
\end{tabular}%
}

\captionof{table}{Candidate model pool. Prices are per 1M tokens.}
\label{tab:model_pool}

\end{minipage}
\end{center}

We select 11 representative VLMs as the candidate pool $\mathcal{M}$,
spanning diverse capability tiers and cost ranges
(Table~\ref{tab:model_pool}), including
Qwen3-VL-8B~\cite{bai2025qwen3},
Qwen3.5-35B-A3B~\cite{qwen35},
Gemini 2.5/3/3.5 Flash series~\cite{gemini25flash,gemini3flash,gemini35flash},
GPT-5.4 mini/5.4~\cite{gpt54,gpt54mini},
Claude Haiku 4.5/Sonnet 4.6/Opus 4.6%
~\cite{claudehaiku45,claudesonnet46,claudeopus46},
and MiniMax M3~\cite{minimaxm3}.
Each model executes the full Execution Context on all samples,
yielding
$Y \in \{0,1\}^{N\times11}$
and
$C \in \mathbb{R}_+^{N\times11}$.

\paragraph{Routing Methods.}
\label{sec:routing_methods}

We evaluate four categories of routing methods (detailed descriptions in Appendix~\ref{app:implementation}):

\noindent\textbf{Training-free baselines.} Oracle is the theoretical upper bound and selects the cheapest correct model; Strongest always selects the globally best-performing model; Cheapest always selects the globally cheapest model.

\noindent\textbf{Feature-level routers.} These methods use embeddings from frozen encoders to make routing decisions via non-parametric methods, clustering, or lightweight classifiers. They include KNN~\cite{fix1951discriminatory}, PRkNN, K-means~\cite{macqueen1967some}, Linear~\cite{mcfadden1974conditional}, MLP~\cite{rumelhart1986learning}, OVR~\cite{DBLP:journals/jmlr/RifkinK03}, and UniRoute. We adopt UniRoute as the open-set baseline because its routing mechanism is modality-agnostic; other open-set routers rely on text-only features and require non-trivial redesign for multimodal settings.

\noindent\textbf{End-to-end routers.} These methods learn the routing decision function end-to-end and include CosineCls, RouterDC, ZOOTER, and VLC.

\noindent\textbf{SCOPE-Router (Ours).} Our method is a calibration-profile-based dual-tower matching router trained with CRM+RCCR (see the Method section).

\paragraph{Evaluation Metrics.}
\label{sec:metrics}

\begin{wrapfigure}{R}{0.48\columnwidth}
    \centering
    \includegraphics[width=\linewidth]{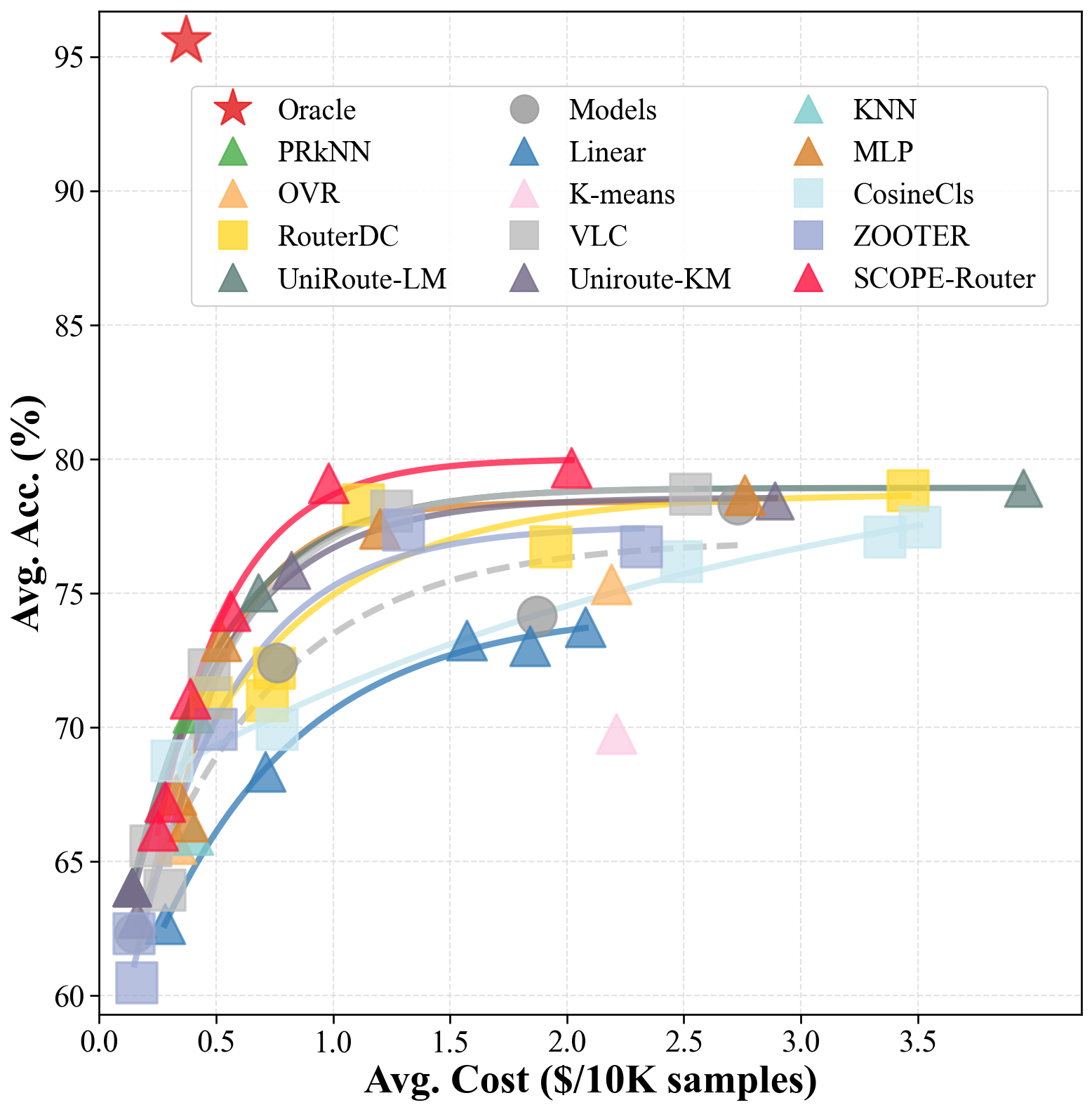}
    \caption{Accuracy--cost trade-offs on VL-RouterBench. Curves trace each router's Pareto front across $\lambda$; points nearer the upper-left are better.}
    \label{fig:pareto}
\end{wrapfigure}

For each router, we report Average Accuracy (fraction of test samples routed correctly), Average Cost (mean inference cost in \$/10K samples), and Rank Score $S_\beta$ (Eq.~\eqref{eq:rank_score}, $\beta = 0.1$). Rank Score serves as the primary ranking criterion. Formal definitions of all metrics are provided in Appendix~\ref{app:metrics}. We also sweep $\lambda$ on VL-RouterBench to trace each router's accuracy--cost Pareto frontier (Figure~\ref{fig:pareto}).

\FloatBarrier
\section{Experiments}

\subsection{Main Results}

\begin{figure*}[t]
    \centering
    \includegraphics[width=0.98\textwidth]{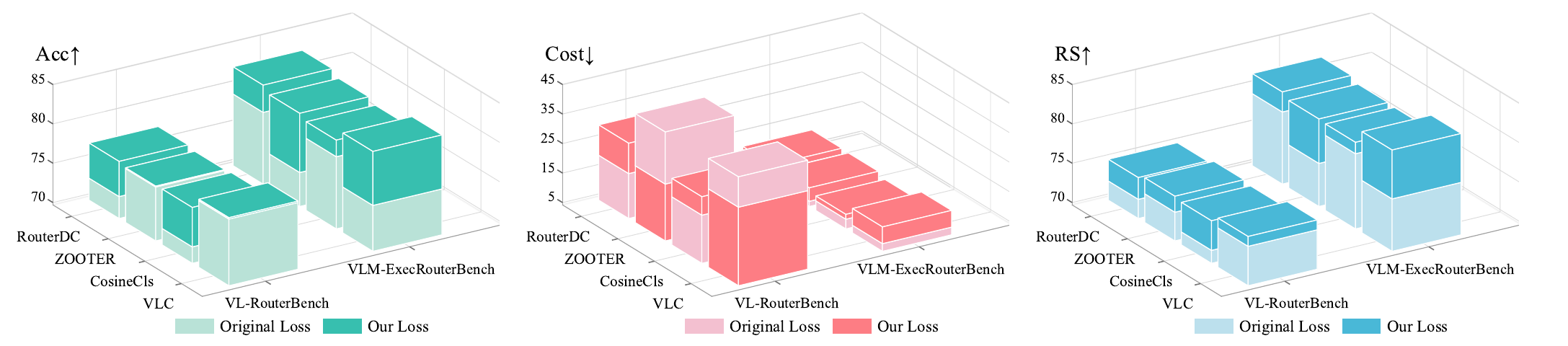}
    \caption{
    Transferability of our cost-aware loss across different end-to-end routers
    on VL-RouterBench and VLM-ExecRouterBench.
    Applying our loss consistently improves the overall routing performance
    across RouterDC, ZOOTER, CosineCls, and VLC.
    }
    \label{fig:loss_transferability}
\end{figure*}

\begin{table*}[!t]
\centering

\scriptsize
\setlength{\tabcolsep}{1.6pt}
\renewcommand{\arraystretch}{0.96}

\begin{adjustbox}{max width=\textwidth}
\begin{tabular}{c|cccc|cccc|cccc}
\toprule

\multirow{2}{*}{\textbf{Router}}
& \multicolumn{4}{c|}{\textbf{VLM-ExecRouterBench}}
& \multicolumn{4}{c|}{\textbf{VL-RouterBench}}
& \multicolumn{4}{c}{\textbf{MMR-Bench}} \\

\cmidrule(lr){2-5}
\cmidrule(lr){6-9}
\cmidrule(lr){10-13}

& \textbf{Avg.\ Acc.} $\uparrow$
& \textbf{Avg.\ Cost} $\downarrow$
& \textbf{Rank Score} $\uparrow$
& \textbf{Rank} $\downarrow$
& \textbf{Avg.\ Acc.} $\uparrow$
& \textbf{Avg.\ Cost} $\downarrow$
& \textbf{Rank Score} $\uparrow$
& \textbf{Rank} $\downarrow$
& \textbf{Avg.\ Acc.} $\uparrow$
& \textbf{Avg.\ Cost} $\downarrow$
& \textbf{Rank Score} $\uparrow$
& \textbf{Rank} $\downarrow$ \\

\midrule
\multicolumn{13}{c}{\textit{Training-free Baselines}} \\

Oracle
& 95.25 & 59.53 & 91.77 & 0
& 95.60 & 0.37 & 93.68 & 0
& 91.67 & 10.41 & 75.67 & 0 \\

Strongest
& \textbf{86.86} & 308.24 & 66.75 & 13
& 78.01 & 2.72 & 68.88 & 11
& \textbf{74.25} & 81.30 & 50.64 & 14 \\

Cheapest
& 67.74 & \textbf{18.94} & 69.78 & 12
& 62.43 & \textbf{0.14} & 64.63 & 14
& 42.25 & \textbf{0.25} & 43.20 & 15 \\

\midrule
\multicolumn{13}{c}{\textit{End-to-End Routers}} \\

CosineCls
& $79.17\pm0.46$
& $33.70\pm3.73$
& \underline{$79.55\pm0.50$}
& \underline{2}
& $70.24\pm0.62$
& $0.61\pm0.16$
& $69.90\pm0.34$
& 9
& $61.31\pm0.58$
& $7.37\pm1.98$
& $56.05\pm0.77$
& 7 \\

RouterDC
& $79.11\pm0.84$
& $38.65\pm5.04$
& $79.16\pm0.84$
& 3
& $77.52\pm1.04$
& $1.04\pm0.26$
& \underline{$74.59\pm1.05$}
& \underline{2}
& $62.22\pm0.88$
& $7.33\pm1.78$
& $56.75\pm0.89$
& 6 \\

ZOOTER
& $74.32\pm1.23$
& \underline{$27.55\pm4.87$}
& $75.47\pm1.20$
& 9
& $74.65\pm1.34$
& $0.93\pm0.15$
& $72.58\pm1.03$
& 7
& $61.65\pm1.21$
& $3.62\pm0.63$
& $57.77\pm1.02$
& 5 \\

VLC
& $75.83\pm1.20$
& $30.93\pm0.80$
& $76.66\pm1.12$
& 6
& \underline{$78.09\pm1.17$}
& $1.23\pm0.03$
& $74.33\pm0.51$
& 3
& $65.06\pm1.03$
& $5.50\pm0.13$
& $59.59\pm0.79$
& 3 \\

\midrule
\multicolumn{13}{c}{\textit{Feature-level Routers}} \\

KNN
& 70.81 & 40.93 & 71.41 & 11
& 66.26 & \underline{0.38} & 67.13 & 12
& 58.66 & 7.07 & 54.11 & 10 \\

PRkNN
& 75.10 & 40.72 & 75.36 & 10
& 70.68 & 0.41 & 71.09 & 8
& 59.72 & 7.30 & 54.85 & 9 \\

OVR
& 85.27 & 407.16 & 55.14 & 15
& 75.49 & 2.18 & 68.92 & 10
& 69.01 & 55.43 & 51.98 & 12 \\

K-means
& \underline{85.69} & 362.20 & 60.73 & 14
& 70.01 & 2.21 & 64.62 & 15
& 70.12 & 61.59 & 51.63 & 13 \\

Linear
& $79.85\pm1.11$
& $101.15\pm10.01$
& $76.15\pm1.07$
& 8
& $68.32\pm7.45$
& $1.32\pm0.33$
& $65.68\pm5.69$
& 13
& \underline{$70.60\pm1.05$}
& $53.76\pm4.89$
& $53.05\pm0.93$
& 11 \\

MLP
& $75.88\pm0.53$
& $28.45\pm3.55$
& $76.87\pm0.55$
& 5
& $77.49\pm0.56$
& $1.13\pm0.13$
& $74.23\pm0.22$
& 4
& $60.88\pm0.41$
& $7.45\pm0.89$
& $55.69\pm0.42$
& 8 \\

UniRoute-KM
& $76.57\pm0.76$
& $48.16\pm13.13$
& $76.26\pm1.03$
& 7
& $75.88\pm0.73$
& $0.82\pm0.21$
& $74.04\pm1.08$
& 5
& $60.88\pm0.61$
& $0.51\pm0.14$
& \underline{$59.72\pm0.60$}
& \underline{2} \\

UniRoute-LM
& $78.55\pm1.61$
& $42.58\pm8.29$
& $78.40\pm1.55$
& 4
& $75.14\pm0.96$
& $0.68\pm0.14$
& $73.86\pm1.04$
& 6
& $58.23\pm1.16$
& \underline{$0.27\pm0.05$}
& $57.91\pm1.05$
& 4 \\

\textbf{SCOPE-Router (Ours)}
& $81.65\pm1.06$
& $46.20\pm11.62$
& $\mathbf{80.94}\pm1.22$
& \textbf{1}
& $\mathbf{79.11}\pm1.02$
& $0.98\pm0.27$
& $\mathbf{76.18}\pm1.44$
& \textbf{1}
& $62.90\pm0.81$
& $0.65\pm0.17$
& $\mathbf{61.23}\pm0.75$
& \textbf{1} \\

\bottomrule
\end{tabular}
\end{adjustbox}
\caption{
Performance of router methods on VLM-ExecRouterBench, VL-RouterBench,
and MMR-Bench.
Avg.\ Acc.\ denotes average accuracy (\%),
Avg.\ Cost denotes average cost (\$/10K samples),
and methods are ranked by Rank Score.
}
\label{tab:exec_results}
\end{table*}

We evaluate all routing methods on three benchmarks: (1) \textbf{VLM-ExecRouterBench} (ours); (2) \textbf{VL-RouterBench}; and (3) \textbf{MMR-Bench}~\cite{mmrbench}. For each benchmark, all routers are trained on its corresponding training split. We sweep the cost-sensitivity coefficient $\lambda \in \{0,\allowbreak 10,\allowbreak 100,\allowbreak 1000,\allowbreak 10000,\allowbreak +\infty\}$ for each method and report results at the $\lambda$ achieving the best validation Rank Score. Training details and per-method hyperparameters are in Appendix~\ref{app:implementation}.

Table~\ref{tab:exec_results} presents results across all three benchmarks. SCOPE-Router ranks first in Rank Score on all three benchmarks, the only method to achieve this across benchmarks spanning traditional VQA, execution-oriented tasks, and multimodal reasoning.

On \textbf{VLM-ExecRouterBench}, SCOPE-Router obtains a Rank Score of $80.94\pm1.22$, followed by CosineCls ($79.55\pm0.50$). Compared with the Strongest baseline, it reduces cost by 85\% while sacrificing only 5.21pp in accuracy. The top end-to-end routers (CosineCls, RouterDC) outperform MLP on execution tasks---a pattern not observed on VL-RouterBench---suggesting that execution domains' higher inter-task heterogeneity benefits more expressive matching. A per-dataset breakdown in Appendix Table~\ref{tab:vlmexec_per_dataset_appendix} reveals that SCOPE-Router's overall advantage stems from consistent competitiveness across all three domains rather than dominance in any single one: it achieves the highest accuracy on Search (68.18\%) and DocVQA (97.92\%), while maintaining strong cost efficiency on Code and Agentic tasks where cost-unaware methods (OVR, K-means) achieve higher accuracy but at 3--8$\times$ the cost.

On \textbf{VL-RouterBench}, SCOPE-Router scores $76.18\pm1.44$ vs.\ the runner-up RouterDC ($74.59\pm1.05$). The mean$\pm$std intervals overlap, so the ranking difference on this benchmark alone is not conclusive; however, SCOPE-Router simultaneously achieves the highest accuracy ($79.11\pm1.02\%$) and reduces cost by 64\% vs.\ Strongest, indicating a favorable quality--cost tradeoff. Here MLP ($74.23\pm0.22$) performs comparably to end-to-end routers (RouterDC $74.59$, VLC $74.33$), indicating that frozen encoders already provide strong discriminative features for traditional VQA tasks.

On \textbf{MMR-Bench}, SCOPE-Router achieves $61.23\pm0.75$, surpassing UniRoute-KM ($59.72\pm0.60$) by 1.51 points with mean$\pm$std intervals that do not overlap, at merely 0.8\% of the Strongest baseline's cost ($0.65$ vs.\ $81.30$).

\subsection{Ablation Studies}

\paragraph{Loss Function Ablation.}
We ablate the training objective on VL-RouterBench (Table~\ref{tab:loss_ablation}). CRM alone (w/o RCCR) achieves 75.50, a 1.0-point gain over pure softmax (74.51), confirming that independent per-model scoring avoids multi-positive dilution. RCCR is effective only under CRM: adding RCCR to softmax hurts performance (74.51 $\to$ 73.41), whereas adding it to CRM improves Rank Score from 75.50 to 76.18 (+0.68). This reveals that CRM's unnormalized scoring framework is a prerequisite for RCCR's distributional regularization to be beneficial---the two components are complementary rather than independently additive.

\begin{table}[!htbp]
\centering
\scriptsize
\setlength{\tabcolsep}{2.4pt}
\renewcommand{\arraystretch}{0.92}

\begin{adjustbox}{max width=\columnwidth}
\begin{tabular}{l|ccc}
\toprule
\textbf{Configuration} & \textbf{Avg.\ Acc.} $\uparrow$ & \textbf{Avg.\ Cost} $\downarrow$ & \textbf{Rank Score} $\uparrow$ \\
\midrule
\textbf{Full model} & \textbf{79.11} & 0.98 & \textbf{76.18} \\
w/o RCCR & 77.90 & 0.89 & 75.50 \\
w/o CRM & 75.82 & 0.97 & 73.41 \\
w/o CRM, RCCR & 76.13 & 0.76 & 74.51 \\
\bottomrule
\end{tabular}
\end{adjustbox}
\caption{
Loss function ablation on VL-RouterBench.
RCCR denotes Routing-Consistency Contrastive Regularization.
}
\label{tab:loss_ablation}
\end{table}

\paragraph{Generalizability of the Loss Function.}
CRM+RCCR applied as a replacement loss to RouterDC, ZOOTER, CosineCls, and VLC improves Rank Score on both benchmarks (Figure~\ref{fig:loss_transferability}), with gains ranging from +1.25 to +6.21 points. Gains are larger on VLM-ExecRouterBench, where execution tasks' stronger inter-model complementarity amplifies the benefit of eliminating multi-positive dilution. This confirms the objective's cross-architecture generalizability.

\paragraph{OOD Generalization.}
\label{sec:ood}
We select 5 datasets from VLM-ExecRouterBench (MBPP, LiveCodeBench, ChartQA, OCRBench, RealWorldQA) as OOD test sets, with the remaining data split 9:1 into training and validation sets (Table~\ref{tab:ood}). All 11 models remain in the candidate pool; only the task distribution shifts. SCOPE-Router achieves the highest Rank Score (88.14), outperforming the runner-up K-means (86.30) by 1.84 points at only 21.8\% of its cost. Its OOD score exceeds the in-distribution score (80.94), consistent with higher Oracle performance in the OOD set (97.15 vs.\ 91.77), indicating stronger inter-model complementarity in these held-out tasks.

\begin{table*}[!t]
\centering

\scriptsize
\setlength{\tabcolsep}{3.2pt}
\renewcommand{\arraystretch}{0.96}

\begin{adjustbox}{max width=\textwidth}
\begin{tabular}{l|cccc|cccc}
\toprule

\multirow{2}{*}{\textbf{Router}}
& \multicolumn{4}{c|}{\textbf{In-Distribution}}
& \multicolumn{4}{c}{\textbf{Out-of-Distribution}} \\

\cmidrule(lr){2-5}
\cmidrule(lr){6-9}

& \textbf{Avg.\ Acc.} $\uparrow$
& \textbf{Avg.\ Cost} $\downarrow$
& \textbf{Rank Score} $\uparrow$
& \textbf{Rank} $\downarrow$
& \textbf{Avg.\ Acc.} $\uparrow$
& \textbf{Avg.\ Cost} $\downarrow$
& \textbf{Rank Score} $\uparrow$
& \textbf{Rank} $\downarrow$ \\

\midrule
UniRoute-KM & 73.58 & 31.78 & 74.51 & 2 & 76.95 & 3.21 & 78.60 & 2 \\
UniRoute-LM & 73.05 & 31.03 & 74.06 & 3 & 75.86 & 2.82 & 77.56 & 3 \\
\textbf{SCOPE-Router (Ours)} & \textbf{78.44} & 82.93 & \textbf{75.96} & \textbf{1} & \textbf{84.21} & 19.84 & \textbf{85.35} & \textbf{1} \\
\bottomrule

\end{tabular}
\end{adjustbox}
\caption{
Open-set routing results with 5 of 11 candidate models held out
under in-distribution and out-of-distribution settings.
}
\label{tab:open_set_id}
\end{table*}

\paragraph{Open-Set Routing.}
We conduct open-set experiments on VLM-ExecRouterBench by holding out 5 of 11 candidate models as unseen (Gemini 2.5 Flash Lite, MiniMax M3, GPT-5.4, Gemini 3.5 Flash, Claude Opus 4.6), excluded from profile construction and router training. At evaluation time, their profiles are generated from calibration samples and added to the candidate pool. On in-distribution data, SCOPE-Router achieves Rank Score 75.96 (Table~\ref{tab:open_set_id}), outperforming UniRoute-KM by 1.45 points. Under doubly OOD evaluation---unseen models on the same 5 held-out datasets (Table~\ref{tab:open_set_id})---SCOPE-Router's advantage expands to +6.75 Rank Score and +7.26 accuracy over UniRoute-KM, demonstrating that profile-based matching is more robust than cluster-based methods under distribution shifts.

\paragraph{MLP Hidden Dimension.}
The optimum is 128 dimensions (Table~\ref{tab:hidden_dim}), though scores span only 0.62 points across \{64--4096\}, indicating low sensitivity.

\paragraph{Freezing vs.\ Unfreezing the Query Embedder.}
Freezing the query embedder outperforms unfreezing on VL-RouterBench (76.18 vs.\ 75.68; Table~\ref{tab:freeze}) while reducing training cost; we adopt freezing as default.

\paragraph{Calibration Strategy Ablation.}
Among single strategies (Table~\ref{tab:calib_ablation}), Diversity achieves the best Rank Score (75.69); the full triple combination achieves 76.18, confirming complementarity.

\paragraph{Text and Vision Encoder Selection.}
Across 25 encoder combinations (Figure~\ref{fig:encoder_ablation}), Rank Scores span $<$0.8 points (75.54--76.33), confirming robustness to encoder choice.

\FloatBarrier
\section{Conclusion}

We present VLM-ExecRouterBench, SCOPE-Router, and CRM+RCCR, advancing VLM routing along evaluation coverage, open-set capability, and cost-aware training. SCOPE-Router achieves the best Rank Score on all three benchmarks, leads by 1.84 points under OOD and 6.75 points under doubly OOD open-set evaluation. CRM+RCCR consistently improves four architecturally diverse routers by +1.25 to +6.21 points. Limitations are discussed in Appendix~\ref{app:limitations}.
\setcitestyle{numbers,square}
\bibliography{citation}

\newpage
\EnableTOC
\clearpage
\appendix

\section*{Appendix}
\begingroup
\setcounter{tocdepth}{2}  
\tableofcontents
\endgroup

\setcounter{table}{0}
\setcounter{figure}{0}
\section{Data Generation}
\label{app:data_generation}

VLM-ExecRouterBench provides a reproducible routing benchmark built from
standardized executions of a shared candidate VLM pool. The final benchmark
contains 33,966 samples and 373,626 sample--model executions, covering 12
datasets and 11 candidate VLMs. To support reuse and independent extension, we
will release the data-generation pipeline together with the benchmark data,
including the source conversion scripts, executor prompts, tool schemas,
verification code, and saved execution trajectories. Figure~\ref{fig:data_generation_overview}
summarizes the end-to-end construction process across the three execution
domains.

\paragraph{Unified task interface.}
Each source example is first converted into a common execution-oriented schema
with three components. The \textbf{Routing Input} contains only the
pre-execution information available to the router. The \textbf{Execution
Context} contains the complete prompt, visual inputs, tool specification, or
retrieval context sent to the candidate VLM. The \textbf{Verification Rule}
specifies the task-dependent checker used to map a model response to a binary
correctness label. This design keeps the router input lightweight while
evaluating each VLM under the full execution protocol.

\paragraph{Execution backends and tools.}
The three benchmark domains share the same schema but use different execution
backends.

\textbf{Code.} Code examples are converted into a unified executor prompt containing the problem statement and available public tests. Candidate VLMs return code solutions, which are checked by unit tests, input--output tests, or official benchmark evaluators depending on the source dataset.

\textbf{Agentic.} Multimodal document and visual reasoning examples are executed
through a lightweight multimodal agent loop. The agent receives the image and
question directly, can call image tools before answering, and must return a final
natural-language response. The available tools include image inspection for
dimensions, crop and zoom operations for isolating local visual regions, and an
OCR\cite{zhang2026ppocrv6} tool for reading image text. OCR is implemented with a PaddleOCR-backed HTTP
service that returns recognized text blocks, confidence scores, and bounding
boxes. This tool interface makes document, chart, diagram, and OCR-heavy VQA
tasks closer to realistic interactive VLM use, rather than a single opaque
image-to-answer call.

\textbf{Search.} BrowseComp-Plus examples are executed with a bounded
fixed-corpus retrieval tool. Each model may issue retrieval calls against the
provided corpus and then must return an exact final answer string. Retrieval is
implemented with a dense retriever based on Qwen3-Embedding-8B\cite{zhang2025qwen3embedding}: query embeddings
are generated with the retrieval instruction prefix used for web-search-style
passage retrieval, L2-normalized, and scored against precomputed corpus
embeddings by inner product. Each retrieval call returns the top five passages
with their ranks, similarity scores, document identifiers, and available fields
such as title and URL. Passage text is capped at 2,400 characters before being
returned to the model, which keeps the search context bounded while still
preserving enough evidence for multi-hop answer synthesis.

\paragraph{Trajectory logging.}
In addition to final correctness labels, the generation pipeline records
execution trajectories for downstream analysis. For multimodal agent tasks, the
trajectory contains the model turns, tool schemas, issued tool calls, crop/zoom
or OCR results, tool-budget state, and final answer. For Search tasks, the
trajectory records retrieval queries, retrieved document identifiers, snippets,
and final answers. These traces are not used by the router at inference time,
but they make the benchmark useful for studying model tool-use behavior,
failure modes, and future trajectory-aware routing methods.

\paragraph{Verification and matrix construction.}
Responses are verified using the strongest available automatic checker for each
source: executable tests for code, normalized exact/choice matching for
short-answer visual tasks, GPT-5.5\cite{openai2026gpt55} semantic judging only when rule-based
matching is insufficient, and answer matching against the BrowseComp-Plus
references for search. After verification, every sample--model pair contributes
one entry to the correctness matrix $Y \in \{0,1\}^{N \times K}$. Logged token
usage and the corresponding input/output prices define the cost matrix
$C \in \mathbf{R}_{+}^{N \times K}$. The official train/dev/test split contains
23,776/3,396/6,794 samples; the full dataset composition is visualized in the
main paper.

\section{Source Datasets}
\label{app:source_datasets}

\subsection{Code}
\begin{itemize}[leftmargin=*]
\item \textbf{MBPP} contributes 974 entry-level Python programming tasks focused
on basic algorithmic and functional-programming skills. Each problem is paired
with unit tests, making it a clean source for evaluating whether a routed model
can produce executable and semantically correct code rather than merely plausible
text.

\item \textbf{BigCodeBench} contributes 1,139 more realistic Python programming
tasks with richer function specifications and hidden test suites. Compared with
MBPP, it introduces broader library usage, longer prompts, and more
implementation-heavy requirements, making it useful for distinguishing models
that differ in practical coding ability.

\item \textbf{APPS} contributes 4,507 competitive-programming problems described
by natural-language statements and evaluated through input--output tests. It
emphasizes algorithmic reasoning under stdin/stdout-style execution, where
solutions must generalize beyond visible examples and handle edge cases.

\item \textbf{LiveCodeBench} contributes 400 recent
programming-contest-style problems, covering both function-call and
standard-input formats. Its temporally newer tasks reduce overlap with older
training distributions and provide a sharper test of current code-generation
robustness.
\end{itemize}

\subsection{Agentic}
\begin{itemize}[leftmargin=*]
\item \textbf{MMMU} contributes 10,813 multimodal questions spanning
college-level disciplines and heterogeneous visual formats such as diagrams,
tables, charts, and scientific figures. It is included to evaluate expert-level
visual reasoning that requires both domain knowledge and deliberate multimodal
understanding.

\item \textbf{DocVQA} contributes 5,347 document-image question-answering
examples. It targets visually grounded reading, layout understanding, and
information extraction from scanned or rendered documents, covering a capability
that differs substantially from natural-image recognition.

\item \textbf{ChartQA} contributes 4,411 questions over charts and plots. It
tests whether a model can interpret visual encodings, compare quantities, and
perform chart-specific reasoning, making it a useful diagnostic for models that
can read text but struggle with structured visual data.

\item \textbf{AI2D} contributes 3,049 grade-school science diagram questions.
The dataset stresses diagram parsing, symbolic visual elements, arrows,
object-relation reasoning, and educational science knowledge, complementing
document and chart understanding with diagram-centric reasoning.

\item \textbf{OCRBench} contributes 983 text-centric visual understanding
examples. It evaluates OCR and downstream reasoning over text embedded in images,
capturing failure modes where a model's final answer is bottlenecked by visual
text recognition.

\item \textbf{MathVista} contributes 980 visually grounded mathematical
reasoning problems. It combines figures, tables, diagrams, and quantitative
reasoning, allowing the benchmark to probe fine-grained differences in
multimodal math capability.

\item \textbf{RealWorldQA} contributes 737 questions about real-world images. It
provides open-ended perception and commonsense reasoning over natural scenes,
balancing the more structured document, chart, and diagram datasets in the
Agentic domain.
\end{itemize}

\subsection{Search}
\begin{itemize}[leftmargin=*]
\item \textbf{BrowseComp-Plus} contributes 626 fixed-corpus deep-search
questions. It forms the Search domain: models must retrieve evidence from a
bounded corpus and synthesize the final answer, allowing routing evaluation to
include retrieval-grounded reasoning rather than only direct question answering.
\end{itemize}

\section{Implementation Details}
\label{app:implementation}

\paragraph{Query and profile representation.}
Unless otherwise stated, we use BGE-M3\cite{bge-m3} as the text encoder and DINOv2-large\cite{oquab2023dinov2} as
the vision encoder. Text and vision embeddings are individually normalized and
then concatenated. SCOPE-Router builds a calibration-derived behavior profile
for each candidate VLM. A QueryMLP projects the incoming query representation to
a 64-dimensional routing vector, while a ProfileMLP projects each model profile
to the same 64-dimensional space; the router scores their match in this shared
routing space. This keeps the router independent of model identity embeddings
and allows newly onboarded VLMs to be introduced through calibration profiles.

\paragraph{Calibration.}
The default calibration budget is 1024 training samples. Hybrid calibration
combines random coverage, diagnostic samples selected by model disagreement and
cost-sensitive utility, and diversity samples selected in the frozen embedding
space. The default mixture uses 50\% random, 30\% diagnostic, and 20\% diversity
samples. Profiles are built from calibration correctness, cost, value scores,
and aggregated task-level behavior statistics.

\paragraph{Training.}
SCOPE-Router maps each query representation and each calibration-derived model
profile into a shared 64-dimensional routing space. The default QueryMLP and
ProfileMLP use 128-dimensional hidden layers with dropout 0.5. The router uses a
dot-product scorer and cost-aware supervision with $\lambda=10$. We optimize
with AdamW using a fixed learning rate of 0.001, weight decay 0.003, and batch
size 512. Training runs for at most 100 epochs
with early stopping patience 20, selecting checkpoints by development Rank
Score. In the main setting, the query encoder is frozen, so training updates the
QueryMLP and ProfileMLP. We use this configuration for the main
experiments unless an ablation explicitly changes the hidden dimension,
calibration strategy, or encoder update policy.

\paragraph{Open-set evaluation.}
For open-set model experiments, held-out VLMs are removed from router training
and are also excluded from calibration-set selection. They are allowed to join
the candidate pool only after behavior profiles are constructed on the fixed
calibration set. This models the practical deployment setting
where a newly released VLM is profiled once on the calibration set before
serving traffic. In strict unseen-pool evaluation, the test-time candidate set
contains only held-out VLMs, so the router must choose among models whose
training labels were never observed.

\section{Evaluation Metrics}
\label{app:metrics}

\paragraph{Accuracy and cost.}
Average accuracy is the fraction of routed samples answered correctly by the
selected VLM. Average cost is computed from the token-level execution cost of the
selected VLM and reported in dollars per 10K samples in the result tables.

\paragraph{Rank Score.}
Following VL-RouterBench, Rank Score harmonically averages accuracy and
log-normalized cost. Given the full-sample cost range $[c_{\min}, c_{\max}]$
and router average cost $\bar{C}$, the normalized cost score is
\begin{equation}
C_{\mathrm{norm}} =
100 \times
\frac{\log_2(c_{\max}) - \log_2(\bar{C})}
{\log_2(c_{\max}) - \log_2(c_{\min})}.
\end{equation}
The factor of 100 aligns the cost term with the percentage scale of average
accuracy. We then combine average accuracy $\bar{A}$ and normalized cost using
the weighted harmonic mean:
\begin{equation}
S(\beta) =
\frac{(1+\beta) \cdot \bar{A} \cdot C_{\mathrm{norm}}}
{\beta \cdot \bar{A} + C_{\mathrm{norm}}},
\end{equation}
where $\beta>0$ controls the preference between quality and cost. We set
$\beta=0.1$ by default to prioritize accuracy. Oracle selects the cheapest
correct model per sample; Strongest always selects the globally most accurate
model; Cheapest always selects the globally cheapest model.

\section{Additional Results}
\label{app:additional_results}

\begin{figure*}[!t]
\centering
\includegraphics[width=0.95\textwidth]{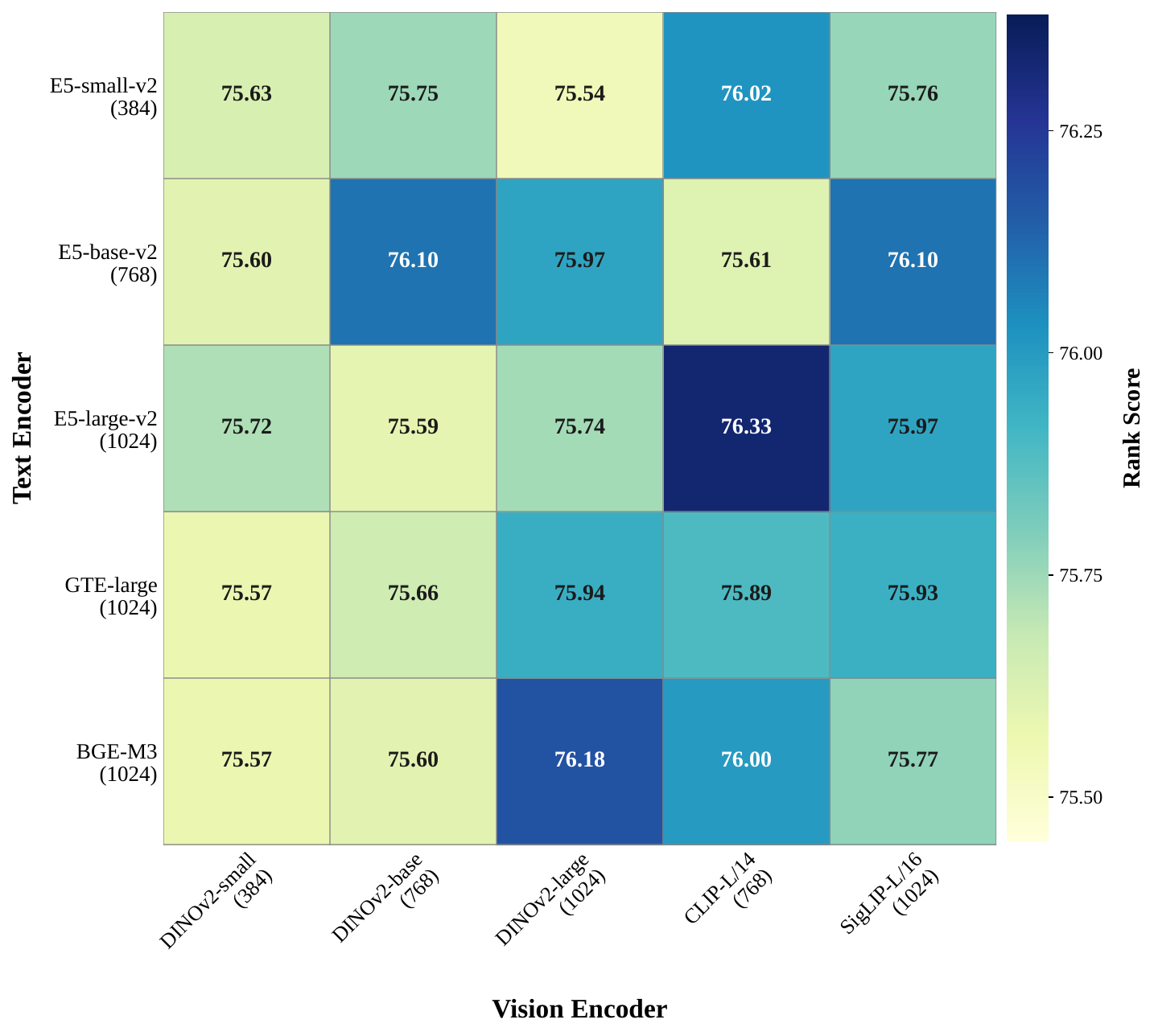}
\caption{
Text and vision encoder selection across 25 encoder combinations.
Rank Scores span less than 0.8 points, demonstrating robustness to
encoder choice.
}
\label{fig:encoder_ablation}
\end{figure*}

\begin{figure*}[!t]
\centering
\includegraphics[width=0.98\textwidth]{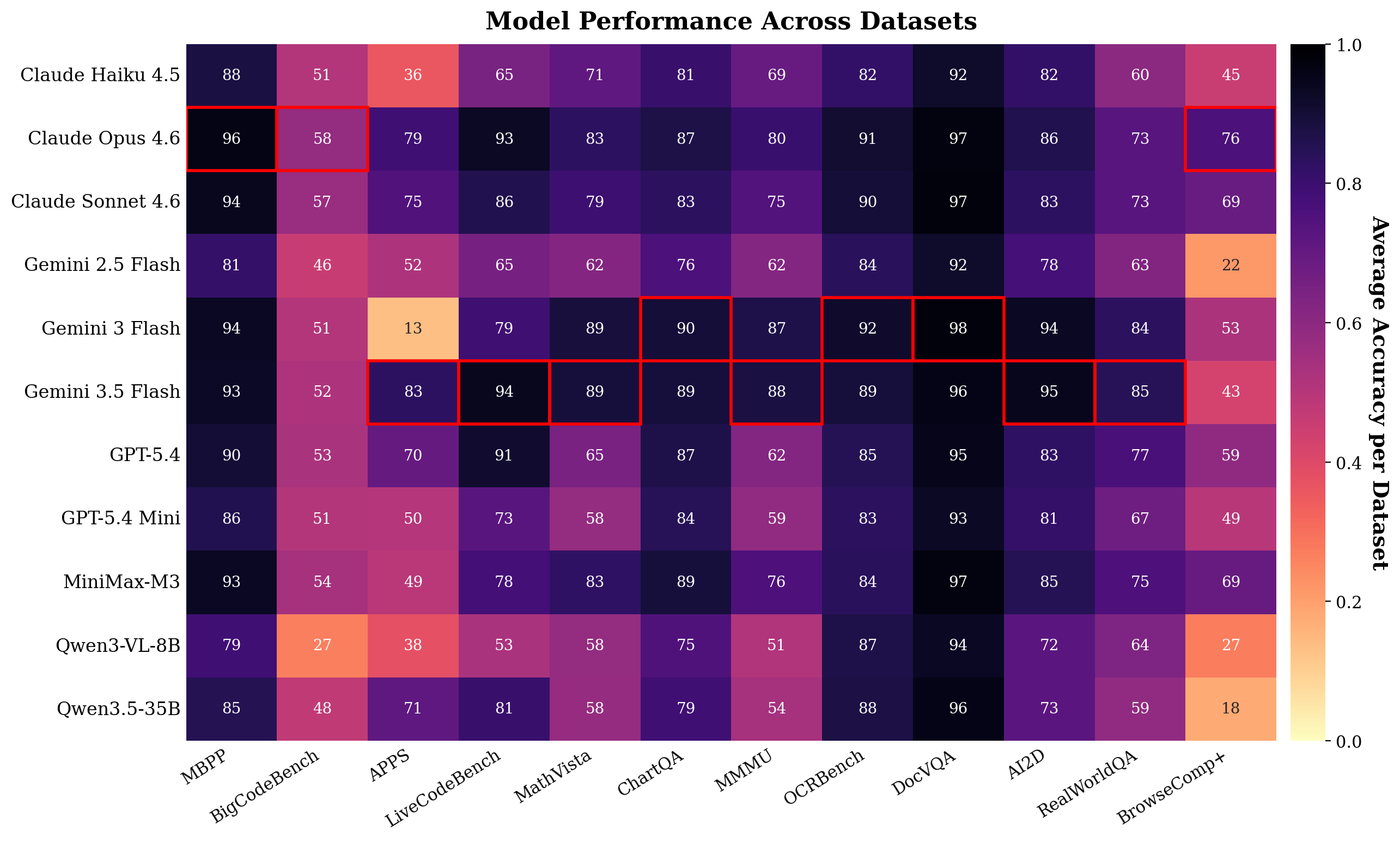}
\caption{
Per-dataset accuracy of each candidate model on VLM-ExecRouterBench.
The heatmap exposes complementary model strengths across code, multimodal
visual reasoning, and search-style datasets.
}
\label{fig:model_dataset_heatmap}
\end{figure*}

\begin{figure*}[t]
\centering
\includegraphics[width=0.88\textwidth]{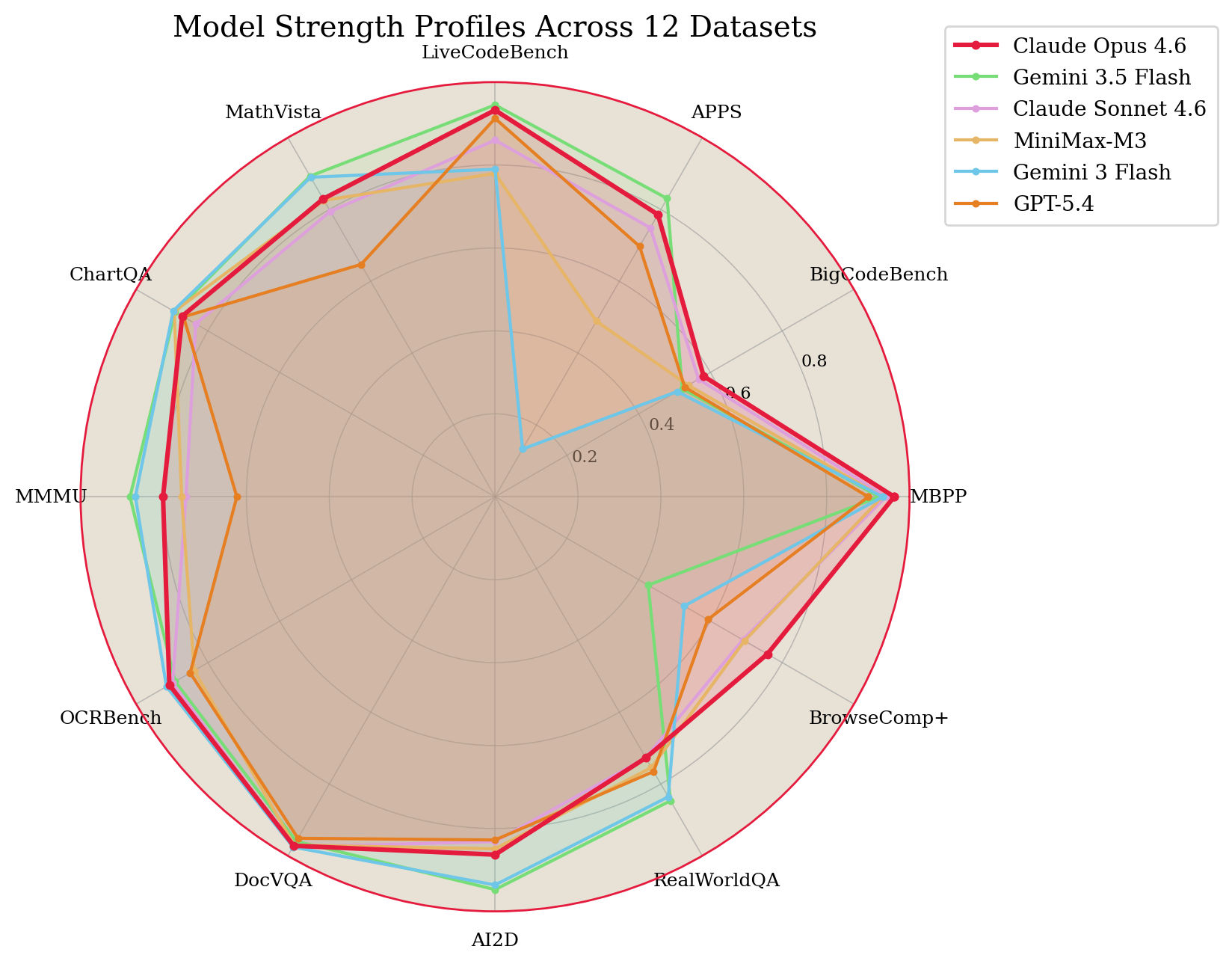}
\caption{
Model strength profiles aggregated by dataset group. The radar plot highlights
that candidate VLMs specialize differently across code, multimodal visual
reasoning, and search-style tasks.
}
\label{fig:model_dataset_radar}
\end{figure*}

\begin{figure*}[t]
\centering
\includegraphics[width=0.95\textwidth]{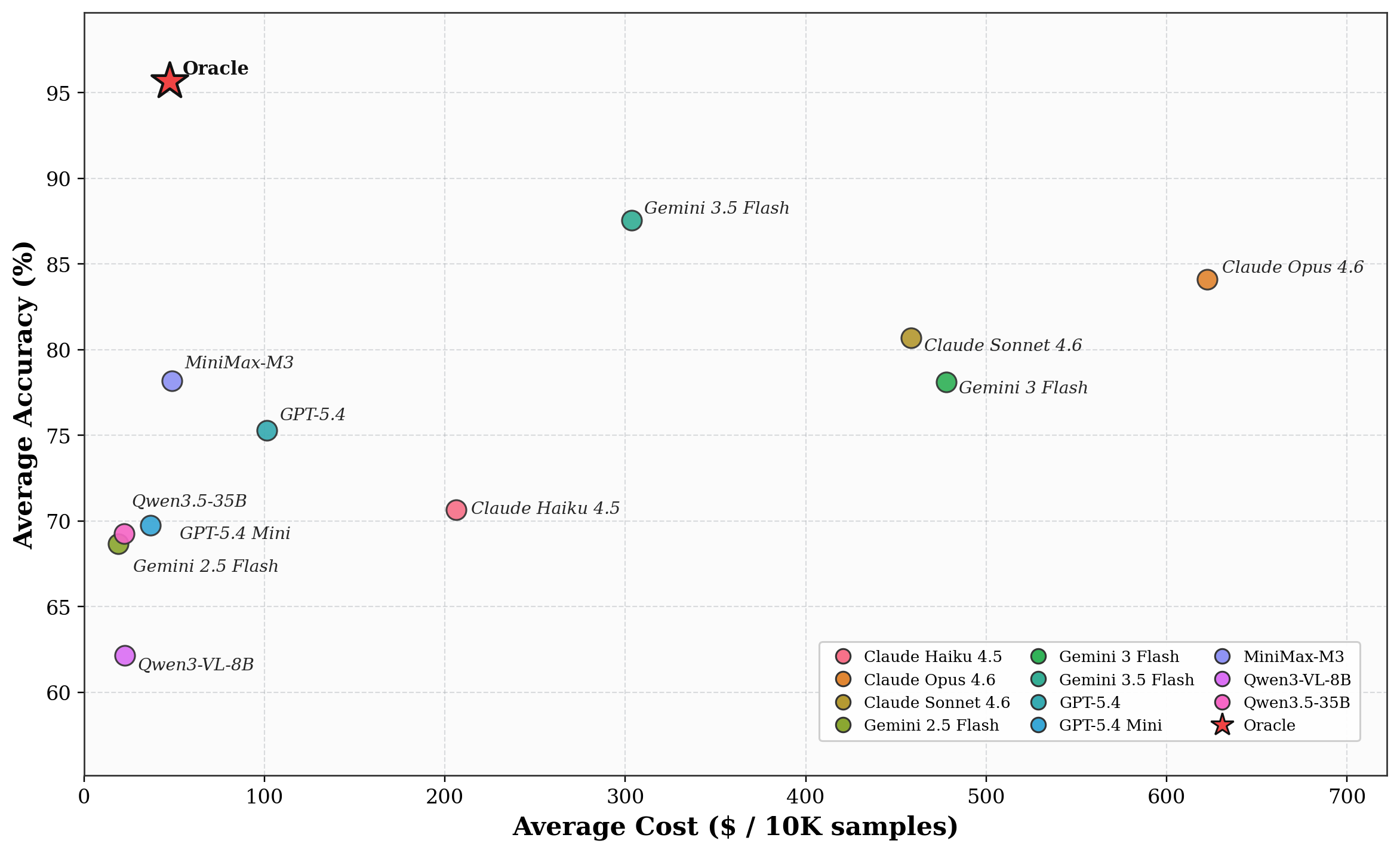}
\caption{
Accuracy--cost distribution of individual candidate VLMs and the oracle
reference computed from the final routing matrix. The plot illustrates that
higher global accuracy often comes with substantially higher execution cost,
motivating cost-aware routing.
}
\label{fig:accuracy_cost_tradeoff}
\end{figure*}

Tables~\ref{tab:ood}--\ref{tab:vlmexec_per_dataset_appendix} and
Figures~\ref{fig:encoder_ablation}--\ref{fig:accuracy_cost_tradeoff}
provide the appendix results referenced in the main paper and the additional
model-level diagnostics. The additional results support four main observations.

First, Table~\ref{tab:ood} shows that SCOPE-Router remains robust on
VLM-ExecRouterBench under the same OOD dataset split and evaluation protocol
used in the main paper, where held-out datasets are used only for testing. Although several
feature-level routers can obtain high accuracy by selecting expensive models,
SCOPE-Router achieves the best Rank Score among learned routers by preserving
high accuracy while avoiding the cost inflation of OVR and K-means. This is the
intended behavior of profile-based open-set routing: the router uses model
profiles to identify useful specialists without collapsing to the strongest or
most expensive model.

\begin{table}[!htbp]
\centering
\caption{
OOD generalization results with all 11 candidate models available.
Rank is determined by Rank Score.
}
\label{tab:ood}

\setlength{\tabcolsep}{2.5pt}
\renewcommand{\arraystretch}{0.96}

\begin{adjustbox}{width=0.6\columnwidth}
\begin{tabular}{c|cccc}
\toprule
\textbf{Router} & \textbf{Avg.\ Acc.} $\uparrow$ & \textbf{Avg.\ Cost} $\downarrow$ & \textbf{Rank Score} $\uparrow$ & \textbf{Rank} $\downarrow$ \\
\midrule

\multicolumn{5}{c}{\textit{Training-free Baselines}} \\
Oracle & 96.87 & 7.68 & 97.15 & 0 \\
Strongest & 89.45 & 224.98 & 75.29 & 15 \\
Cheapest & 75.86 & 2.82 & 77.56 & 14 \\

\midrule
\multicolumn{5}{c}{\textit{End-to-End Routers}} \\
CosineCls & 78.13 & 5.07 & 79.72 & 9 \\
RouterDC & 78.53 & 5.23 & 80.10 & 7 \\
ZOOTER & 77.42 & 5.32 & 79.04 & 11 \\
VLC & 77.31 & 9.24 & 78.94 & 12 \\

\midrule
\multicolumn{5}{c}{\textit{Feature-level Routers}} \\
KNN & 76.63 & 4.34 & 78.29 & 13 \\
PRkNN & 81.31 & 6.06 & 82.71 & 4 \\
OVR & 87.37 & 76.06 & 83.93 & 3 \\
K-means & 88.74 & 58.67 & 86.30 & 2 \\
Linear & 82.37 & 58.97 & 80.76 & 5 \\
MLP & 78.31 & 5.09 & 79.88 & 8 \\
UniRoute-KM & 78.07 & 13.09 & 79.66 & 10 \\
UniRoute-LM & 78.88 & 12.04 & 80.42 & 6 \\
\textbf{SCOPE-Router (Ours)} & \textbf{87.10} & 12.79 & \textbf{88.14} & \textbf{1} \\

\bottomrule
\end{tabular}
\end{adjustbox}
\end{table}

Second, Tables~\ref{tab:hidden_dim}--\ref{tab:calib_ablation} indicate that the
method is not highly sensitive to oversized routing networks, but it does depend
on the quality of the calibration set. A hidden dimension of 128 gives the best
overall tradeoff in our VL-RouterBench ablation, while larger hidden dimensions
do not yield consistent gains. Freezing the query embedder performs on par with
fine-tuning, suggesting that most of the routing signal comes from aligning
frozen query embeddings with model behavior profiles rather than from adapting
the encoder itself. For calibration, the full hybrid strategy combining random,
diagnostic, and diversity sampling outperforms each single-source or two-source
variant, supporting the use of a calibration set that covers both representative
queries and discriminative model-failure cases.

\begin{table}[!htbp]
\centering
\caption{
Effect of MLP hidden dimension on VL-RouterBench.
}
\label{tab:hidden_dim}

\setlength{\tabcolsep}{4pt}
\renewcommand{\arraystretch}{1.0}

\begin{adjustbox}{width=0.6\columnwidth}
\begin{tabular}{c|ccc}
\toprule
\textbf{Hidden Dim} & \textbf{Avg.\ Acc.} $\uparrow$ & \textbf{Avg.\ Cost} $\downarrow$ & \textbf{Rank Score} $\uparrow$ \\
\midrule
64 & 77.37 & 0.77 & 75.56 \\
\textbf{128} & \textbf{79.11} & 0.98 & \textbf{76.18} \\
256 & 78.23 & 0.82 & 76.08 \\
512 & 78.27 & 0.87 & 75.90 \\
1024 & 78.05 & 0.82 & 75.91 \\
2048 & 78.52 & 0.88 & 76.08 \\
4096 & 78.68 & 0.97 & 75.83 \\
\bottomrule
\end{tabular}
\end{adjustbox}
\end{table}

\begin{table}[!htbp]
\centering
\caption{
Effect of freezing or fine-tuning the query embedder on
VL-RouterBench and VLM-ExecRouterBench.
The base setting freezes the query embedder and only trains QueryMLP.
}
\label{tab:freeze}

\scriptsize
\setlength{\tabcolsep}{2.2pt}
\renewcommand{\arraystretch}{0.94}

\begin{adjustbox}{max width=0.6\columnwidth}
\begin{tabular}{l|ccc|ccc}
\toprule
\multirow{2}{*}{\textbf{Variant}}
& \multicolumn{3}{c|}{\textbf{VL-RB}}
& \multicolumn{3}{c}{\textbf{VLM-ExecRB}} \\
\cmidrule(lr){2-4}
\cmidrule(lr){5-7}
& \textbf{Acc.} $\uparrow$
& \textbf{Cost} $\downarrow$
& \textbf{Score} $\uparrow$
& \textbf{Acc.} $\uparrow$
& \textbf{Cost} $\downarrow$
& \textbf{Score} $\uparrow$ \\
\midrule
\textbf{Frozen} & \textbf{79.11} & 0.98 & \textbf{76.18} & 81.65 & 46.20 & 80.94 \\
Fine-tuned & 78.36 & 0.94 & 75.68 & \textbf{81.76} & 44.85 & \textbf{81.14} \\
\bottomrule
\end{tabular}
\end{adjustbox}
\end{table}

\begin{table}[!htbp]
\centering
\caption{
Calibration-strategy ablation on VL-RouterBench.
R, D, and Div denote Random, Diagnostic, and Diversity sampling,
respectively.
}
\label{tab:calib_ablation}

\scriptsize
\setlength{\tabcolsep}{2.2pt}
\renewcommand{\arraystretch}{0.94}

\begin{adjustbox}{max width=\columnwidth}
\begin{tabular}{l|ccc}
\toprule
\textbf{Variant} & \textbf{Avg.\ Acc.} $\uparrow$ & \textbf{Avg.\ Cost} $\downarrow$ & \textbf{Rank Score} $\uparrow$ \\
\midrule
R only & 77.73 & 0.90 & 75.31 \\
D only & 77.34 & 0.80 & 75.37 \\
Div only & 78.59 & 0.99 & 75.69 \\
R+D & 78.87 & 0.96 & 75.78 \\
R+Div & 78.44 & 0.88 & 76.00 \\
D+Div & 78.37 & 0.90 & 75.87 \\
\textbf{R+D+Div (Full)} & \textbf{79.11} & 0.98 & \textbf{76.18} \\
\bottomrule
\end{tabular}
\end{adjustbox}
\end{table}

Third, Table~\ref{tab:vlmexec_per_dataset_appendix} shows that router behavior
varies substantially across execution domains. Document, chart, OCR, code, and
search tasks favor different model choices and therefore expose different
failure modes for learned routers. SCOPE-Router is especially useful in this
setting because it routes using both query features and calibration-derived
model profiles, rather than learning a closed-set classifier over a fixed model
identity space. This helps explain why methods that are competitive on one
dataset can degrade on another when the cost--accuracy frontier changes.

\begin{figure*}[t]
\centering
\includegraphics[width=0.98\textwidth,keepaspectratio]{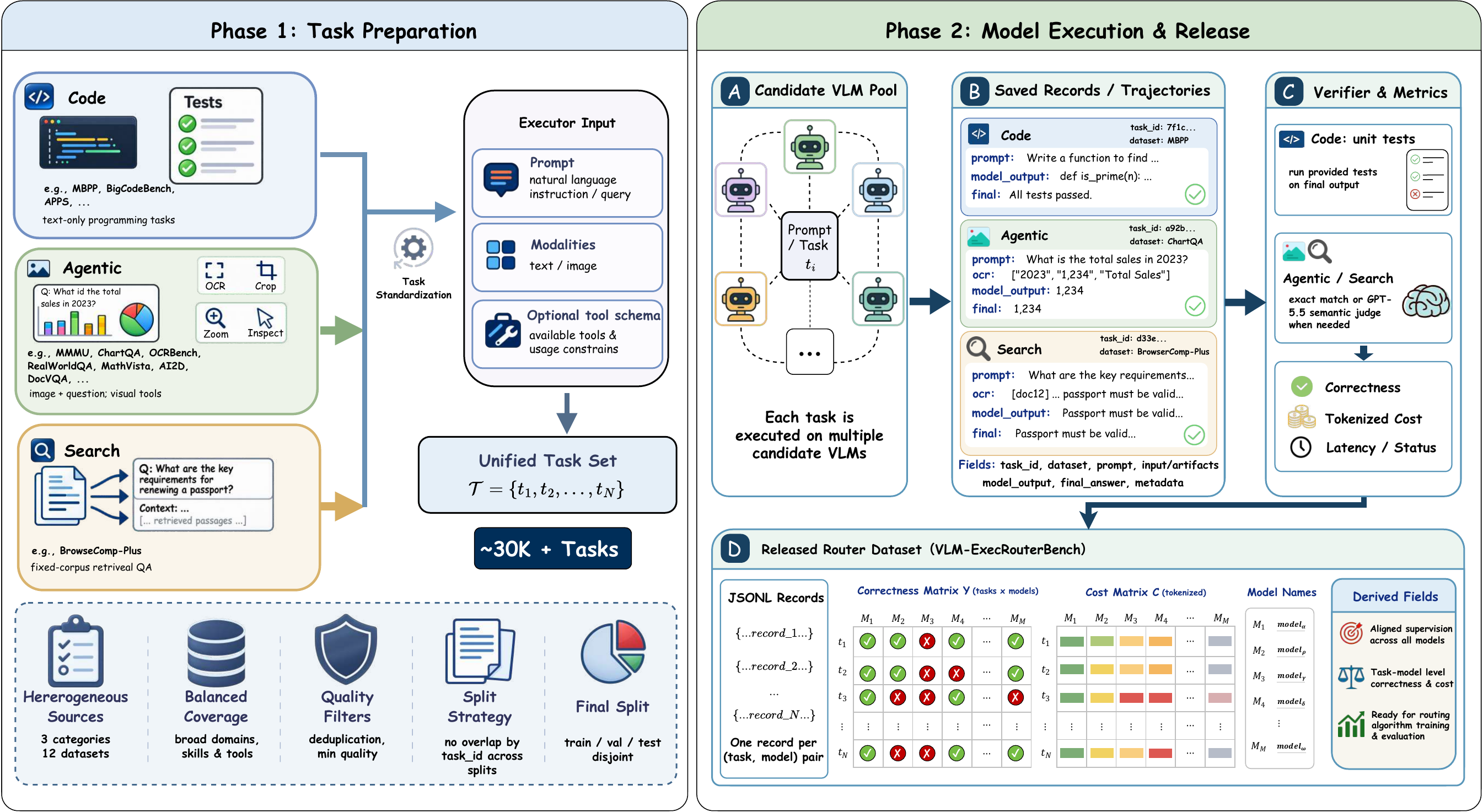}
\caption{
Overview of the VLM-ExecRouterBench data-generation pipeline. Heterogeneous Code, Agentic, and Search source tasks are standardized into executor inputs with the appropriate modalities and optional tool schemas. Candidate VLMs produce responses through the shared executor: Code tasks use one-shot code generation without environment interaction, while Agentic and Search tasks may produce saved tool trajectories. Task-specific verifiers then convert each execution into correctness and cost entries. The resulting JSONL records and aligned correctness/cost matrices form the released router dataset.
}
\label{fig:data_generation_overview}
\end{figure*}

Finally, Figure~\ref{fig:encoder_ablation},
Figure~\ref{fig:model_dataset_heatmap},
Figure~\ref{fig:model_dataset_radar}, and
Figure~\ref{fig:accuracy_cost_tradeoff} provide diagnostics for the
representation and candidate-model pool, while
Figure~\ref{fig:data_generation_overview} complements the earlier
data-generation description with a visual overview of the three execution
domains. The encoder ablation shows that
routing quality depends on the joint text--vision representation. The
per-dataset heatmap further shows that candidate VLMs are not uniformly strong
across datasets. The radar plot summarizes the same heterogeneity at the level
of task families, while the accuracy--cost plot shows that global model strength
and inference cost are not aligned. These visualizations motivate the benchmark
design: open-set routing is meaningful precisely because model strengths are
heterogeneous across task families, and cost-efficient deployment requires
selecting among those strengths rather than ranking models by a single global
accuracy number.

\section{Limitations}
\label{app:limitations}

SCOPE-Router extends naturally to newly added VLMs, but each new model must still be profiled on a labeled calibration set before it can be routed reliably. As a result, open-set performance depends on how well the calibration examples expose the capabilities and failure modes of the incoming model. In addition, the current router makes a single task-level decision and uses the selected model for the entire execution. This matches the benchmark setting, where each sample is represented by one correctness and cost entry per candidate VLM, but it does not adapt to changing requirements within long-horizon Agentic or Search trajectories. For instance, an early step may require fine-grained OCR or visual localization, whereas a later step may require retrieval synthesis or answer normalization. Future work will study more robust model profiling from compact calibration sets and dynamic routing policies that can revise model choices across intermediate execution steps.

\begin{table*}[t]
\centering
\caption{Performance comparison of router methods on VLM-ExecRouterBench across datasets. Avg. Acc. is average accuracy (\%), Avg. Cost is average cost (\$/10K samples). The best and second-best results among learned routers are highlighted in \textbf{bold} and \underline{underlined}, respectively, excluding Oracle and baseline methods.}
\label{tab:vlmexec_per_dataset_appendix}

\scriptsize
\setlength{\tabcolsep}{1.4pt}
\renewcommand{\arraystretch}{0.92}

\begin{adjustbox}{max width=\textwidth}
\begin{tabular}{ll l|rrr|rrrr|rrrrrrrrr}
\toprule

\multirow{2}{*}{\textbf{Group}}
& \multirow{2}{*}{\textbf{Dataset}}
& \multirow{2}{*}{\textbf{Metric}}
& \multicolumn{3}{c|}{\textbf{Baselines}}
& \multicolumn{4}{c|}{\textbf{Hard Label}}
& \multicolumn{9}{c}{\textbf{Soft Label ($\lambda=10$)}} \\

\cmidrule(lr){4-6}
\cmidrule(lr){7-10}
\cmidrule(lr){11-19}

& & &
\textbf{Oracle}
& \textbf{Strong.}
& \textbf{Cheap.}
& \textbf{KNN}
& \textbf{PRkNN}
& \textbf{OVR}
& \textbf{K-means}
& \textbf{Linear}
& \textbf{MLP}
& \textbf{CosCls}
& \textbf{RDC}
& \textbf{VLC}
& \textbf{Zoot.}
& \textbf{U-KM}
& \textbf{U-LM}
& \textbf{SCOPE} \\

\midrule

\multirow{16}{*}{\rotatebox{90}{\textit{Agentic}}}
& \multirow{2}{*}{DocVQA}
& Avg. Acc. $\uparrow$
& 99.53 & 96.22 & 92.62
& 93.00 & 94.13 & \underline{97.54} & 93.66
& 94.42 & 93.19 & 94.04 & 93.66 & 93.09 & 92.81 & 94.32 & 95.84 & \textbf{97.92} \\

& & Avg. Cost $\downarrow$
& 3.77 & 246.03 & 3.01
& 3.49 & 5.13 & 86.96 & 38.03
& 57.08 & \underline{3.40} & 4.31 & 4.82 & 3.67 & \textbf{3.15} & 26.77 & 39.99 & 14.31 \\

\cmidrule(l){2-19}

& \multirow{2}{*}{OCRBench}
& Avg. Acc. $\uparrow$
& 97.00 & 88.00 & 80.50
& 85.50 & 84.00 & \underline{89.50} & 87.00
& 85.00 & 89.50 & \textbf{91.00} & 88.00 & 88.00 & 86.00 & 81.00 & 84.50 & 82.00 \\

& & Avg. Cost $\downarrow$
& 5.78 & 221.49 & 2.45
& \textbf{2.84} & 3.30 & 81.00 & 26.93
& 79.33 & 4.25 & 3.59 & 3.69 & 3.43 & \underline{3.04} & 7.21 & 10.65 & 5.48 \\

\cmidrule(l){2-19}

& \multirow{2}{*}{ChartQA}
& Avg. Acc. $\uparrow$
& 95.63 & 89.51 & 76.07
& 78.58 & 84.70 & \textbf{88.52} & 87.65
& 80.11 & 84.37 & \underline{87.87} & 87.32 & 86.89 & 83.83 & 80.11 & 83.06 & 87.87 \\

& & Avg. Cost $\downarrow$
& 15.73 & 269.70 & 2.43
& \textbf{3.11} & 3.73 & 65.47 & 25.18
& 32.35 & 3.81 & 4.68 & 4.75 & 4.59 & \underline{3.26} & 4.25 & 11.01 & 5.57 \\

\cmidrule(l){2-19}

& \multirow{2}{*}{AI2D}
& Avg. Acc. $\uparrow$
& 99.84 & 93.74 & 77.37
& 80.42 & 82.99 & \underline{91.81} & \textbf{93.26}
& 89.57 & 82.18 & 85.39 & 85.07 & 85.07 & 84.11 & 82.66 & 82.99 & 85.07 \\

& & Avg. Cost $\downarrow$
& 7.19 & 165.90 & 6.60
& \underline{7.42} & \textbf{6.75} & 101.97 & 93.96
& 68.15 & 7.91 & 8.53 & 8.56 & 8.56 & 7.73 & 11.48 & 8.54 & 8.56 \\

\cmidrule(l){2-19}

& \multirow{2}{*}{MathVista}
& Avg. Acc. $\uparrow$
& 97.52 & 88.61 & 63.86
& 73.27 & 82.18 & \underline{90.10} & \textbf{93.07}
& 83.17 & 82.18 & 81.68 & 81.68 & 81.68 & 80.20 & 77.72 & 82.18 & 82.18 \\

& & Avg. Cost $\downarrow$
& 14.22 & 266.49 & 16.69
& 18.66 & \textbf{12.77} & 170.23 & 159.04
& 108.23 & 16.29 & 15.48 & 15.48 & 15.56 & \underline{15.14} & 16.71 & 15.47 & 17.59 \\

\cmidrule(l){2-19}

& \multirow{2}{*}{MMMU}
& Avg. Acc. $\uparrow$
& 96.06 & 86.71 & 60.91
& 63.41 & 67.72 & \underline{82.40} & \textbf{83.60}
& 75.41 & 66.37 & 73.74 & 74.34 & 73.78 & 72.77 & 70.17 & 72.30 & 74.43 \\

& & Avg. Cost $\downarrow$
& 21.06 & 335.52 & 19.39
& \underline{19.69} & 27.38 & 211.32 & 196.39
& 119.19 & \textbf{18.91} & 35.15 & 34.70 & 33.43 & 28.96 & 32.26 & 34.44 & 38.10 \\

\cmidrule(l){2-19}

& \multirow{2}{*}{RealWorldQA}
& Avg. Acc. $\uparrow$
& 98.59 & 79.58 & 59.15
& 58.45 & 61.27 & 72.54 & \underline{76.06}
& \textbf{79.58} & 64.08 & 70.42 & 71.83 & 61.27 & 61.97 & 71.13 & 69.01 & 71.13 \\

& & Avg. Cost $\downarrow$
& 9.74 & 400.87 & 3.16
& \underline{5.05} & 5.40 & 178.59 & 221.78
& 352.91 & 6.90 & 8.73 & 8.62 & 6.96 & \textbf{3.89} & 13.24 & 12.99 & 13.24 \\

\cmidrule(l){2-19}

& \multirow{2}{*}{Agentic Avg.}
& Avg. Acc. $\uparrow$
& 97.28 & 89.85 & 72.59
& 75.01 & 78.71 & \textbf{87.88} & \underline{87.73}
& 82.45 & 78.10 & 82.47 & 82.43 & 81.73 & 80.50 & 78.90 & 80.86 & 83.18 \\

& & Avg. Cost $\downarrow$
& 13.92 & 281.17 & 10.51
& \underline{11.08} & 14.38 & 141.10 & 116.04
& 90.14 & \textbf{10.94} & 17.95 & 17.89 & 17.06 & 14.70 & 21.84 & 26.26 & 21.58 \\

\midrule

\multirow{10}{*}{\rotatebox{90}{\textit{Code}}}
& \multirow{2}{*}{LiveCodeBench}
& Avg. Acc. $\uparrow$
& 94.37 & 92.96 & 70.42
& 71.83 & 80.28 & \underline{91.55} & \textbf{92.96}
& 87.32 & 84.51 & 84.51 & 85.92 & 70.42 & 70.42 & 85.92 & 87.32 & 90.14 \\

& & Avg. Cost $\downarrow$
& 5.49 & 53.90 & 5.33
& 7.39 & 12.47 & 72.86 & 54.76
& 35.39 & 12.61 & 12.89 & 12.96 & \textbf{4.77} & \underline{5.33} & 12.96 & 15.80 & 30.76 \\

\cmidrule(l){2-19}

& \multirow{2}{*}{MBPP}
& Avg. Acc. $\uparrow$
& 99.51 & 92.68 & 74.63
& 76.10 & 75.61 & \textbf{96.10} & \underline{93.17}
& 86.83 & 82.93 & 74.63 & 76.59 & 74.63 & 79.51 & 84.88 & 86.83 & 87.32 \\

& & Avg. Cost $\downarrow$
& 6.10 & 55.82 & 2.86
& 5.33 & 5.63 & 192.55 & 59.55
& 20.01 & 14.10 & \underline{2.86} & 8.46 & \textbf{2.58} & 5.03 & 15.77 & 16.60 & 31.99 \\

\cmidrule(l){2-19}

& \multirow{2}{*}{APPS}
& Avg. Acc. $\uparrow$
& 89.04 & 82.40 & 49.30
& 53.15 & 61.66 & 77.16 & \textbf{81.93}
& 73.43 & 68.30 & 70.28 & 69.23 & 49.42 & 49.42 & 70.51 & 70.05 & \underline{79.84} \\

& & Avg. Cost $\downarrow$
& 41.31 & 122.09 & 28.09
& 40.35 & 50.15 & 174.52 & 141.40
& 78.41 & 54.52 & 54.82 & 54.22 & \underline{30.90} & \textbf{27.08} & 54.91 & 58.34 & 106.83 \\

\cmidrule(l){2-19}

& \multirow{2}{*}{BigCodeBench}
& Avg. Acc. $\uparrow$
& 74.35 & 53.48 & 47.83
& 45.65 & 47.39 & \textbf{55.65} & \underline{55.65}
& 47.83 & 49.57 & 46.96 & 50.00 & 47.39 & 28.26 & 54.35 & 53.91 & 53.04 \\

& & Avg. Cost $\downarrow$
& 12.21 & 40.93 & 1.61
& 4.73 & 3.41 & 93.10 & 59.73
& 13.95 & 3.79 & 3.79 & \underline{3.36} & 11.50 & \textbf{2.40} & 26.78 & 26.48 & 36.57 \\

\cmidrule(l){2-19}

& \multirow{2}{*}{Code Avg.}
& Avg. Acc. $\uparrow$
& 88.42 & 79.62 & 53.96
& 56.30 & 62.32 & \underline{77.13} & \textbf{79.77}
& 71.85 & 68.18 & 67.74 & 67.96 & 53.96 & 51.47 & 70.75 & 70.75 & 76.98 \\

& & Avg. Cost $\downarrow$
& 29.25 & 94.90 & 18.65
& 27.36 & 33.62 & 158.21 & 110.82
& 56.52 & 37.71 & 36.22 & 36.62 & \underline{22.01} & \textbf{18.47} & 42.10 & 44.48 & 79.78 \\

\midrule

\multirow{2}{*}{\rotatebox[origin=c]{90}{\scriptsize\textit{Search}}}
& \multirow{2}{*}{BrowseComp+}
& Avg. Acc. $\uparrow$
& 84.09 & 41.67 & 15.15
& 52.27 & 62.12 & 64.39 & 65.15
& 58.33 & 65.91 & 65.15 & 61.36 & 65.15 & 62.12 & 43.18 & \underline{66.67} & \textbf{68.18} \\

& & Avg. Cost $\downarrow$
& 2203.19 & 3599.38 & 359.94
& 1379.06 & 1171.33 & 13658.61 & 12839.69
& 1004.43 & \textbf{635.84} & 639.89 & 892.86 & 679.74 & \underline{637.13} & 1167.60 & 678.14 & 687.35 \\

\midrule

\multirow{3}{*}{\rotatebox[origin=c]{90}{\scriptsize\textit{Average}}}
& \multirow{3}{*}{All Datasets}
& Avg. Acc. $\uparrow$
& 95.25 & 86.86 & 67.74
& 70.81 & 75.10 & \underline{85.27} & \textbf{85.69}
& 79.85 & 75.88 & 79.17 & 79.11 & 75.83 & 74.32 & 76.57 & 78.55 & 81.65 \\

& & Avg. Cost $\downarrow$
& 59.53 & 308.24 & 18.93
& 40.93 & 40.72 & 407.16 & 362.20
& 101.16 & \underline{28.45} & 33.71 & 38.65 & 30.93 & \textbf{27.55} & 48.16 & 42.58 & 46.20 \\

& & Rank Score $\uparrow$
& 91.77 & 66.75 & 69.78
& 71.41 & 75.36 & 55.14 & 60.73
& 76.15 & 76.87 & \underline{79.55} & 79.16 & 76.66 & 75.47 & 76.26 & 78.40 & \textbf{80.94} \\

\bottomrule
\end{tabular}
\end{adjustbox}
\end{table*}

\clearpage

\clearpage

\end{document}